\documentclass[lettersize,journal]{IEEEtran}
\usepackage{array}
\usepackage[caption=false,font=normalsize,labelfont=sf,textfont=sf]{subfig}
\usepackage{cite,algorithm}
\usepackage{amsmath,amssymb,amsfonts}
\usepackage{algorithmic,tikz,pgfplots}
\usepackage{mathtools}
\usepackage{graphicx,color}
\usepackage{bm}
\usetikzlibrary{positioning,shapes,arrows,arrows.meta,fit,backgrounds,calc}
\usepackage[utf8]{inputenc}
\usepackage{makecell}
\usepackage{multirow}
\usepackage{adjustbox}
\usepackage{import}
\usepackage{xcolor}
\usepackage{fontawesome5}
\usepackage{textcomp}
\usepackage{booktabs,relsize}
\usepackage{anyfontsize}
\usepackage{pgfplots}
\usepackage{lipsum}
\usepackage{hyperref}

\definecolor{block}{RGB}{144, 238, 144}
\newcommand{\FireTok}{\textcolor{red}{\scriptsize\faFire}}
\newcommand{\IceTok}{\textcolor{blue}{\scriptsize\faSnowflake[regular]}}

\begin{document}

\title{Efficient Multi-View 3D Object Detection \\ by Dynamic Token Selection and Fine-Tuning}

\author{Danish Nazir$^{1,2}$, Antoine Hanna-Asaad$^{1}$, Lucas Görnhardt$^{2}$, Jan Piewek$^{1}$, Thorsten Bagdonat$^{1}$, Tim Fingscheidt$^{2}$ 
\thanks{$^{1}$Danish Nazir, Antoine Hanna-Asaad, Jan Piewek, and Thorsten Bagdonat are with Group Innovation, Volkswagen AG, 38440, Wolfsburg, Germany (email: firstname.lastname@volkswagen.de)}

\thanks{$^{2}$Danish Nazir, Lucas Goernhardt, and Tim Fingscheidt are with the Institute for Communications Technology, Technische Universität Braunschweig, 38106 Braunschweig, Germany (email: firstname.lastname@tu-bs.de)
}
}

\markboth{Journal of \LaTeX\ Class Files,~Vol.~14, No.~8, August~2021}%
{Shell \MakeLowercase{\textit{et al.}}: A Sample Article Using IEEEtran.cls for IEEE Journals}



\maketitle

\begin{abstract}
Existing multi-view three-dimensional (3D) object detection approaches widely adopt large-scale pre-trained vision transformer (ViT)-based foundation models as backbones, being computationally complex. To address this problem, current state-of-the-art (SOTA) \texttt{ToC3D} for efficient multi-view ViT-based 3D object detection employs ego-motion-based relevant token selection. However, there are two key limitations: (1) The fixed layer-individual token selection ratios limit computational efficiency during both training and inference. (2) Full end-to-end retraining of the ViT backbone is required for the multi-view 3D object detection method. In this work, we propose an image token compensator combined with a token selection for ViT backbones to accelerate multi-view 3D object detection. Unlike \texttt{ToC3D}, our approach enables dynamic layer-wise token selection within the ViT backbone. Furthermore, we introduce a parameter-efficient fine-tuning strategy, which trains only the proposed modules, thereby reducing the number of fine-tuned parameters from more than $300$ million (M) to only $1.6$ M. Experiments on the large-scale NuScenes dataset across three multi-view 3D object detection approaches demonstrate that our proposed method decreases computational complexity (GFLOPs) by $48\%$ ... $55\%$, inference latency (on an \texttt{NVIDIA-GV100} GPU) by $9\%$ ... $25\%$, while still improving mean average precision by $1.0\%$ ... $2.8\%$ absolute and NuScenes detection score by $0.4\%$ ... $1.2\%$ absolute compared to so-far SOTA \texttt{ToC3D}.  
\end{abstract}



\section{Introduction}

Three-dimensional (3D) object detection from multiple calibrated camera views (multi-view) is a fundamental component of the environment perception stack in autonomous driving \cite{streampetr,petr,toc3D,tgbc}. Recent multi-view 3D object detection methods can be broadly categorized into two branches: (1) dense birds-eye-view (BEV)-based methods \cite{liftspatshoot,bevformerv1, bevformerv2,BEVdet,BEVDepth,Solofusion,li2024bevnext}, and (2) sparse-query-based methods \cite{detr3d,petr,streampetr,mbev,sparse4d,sparsebev,raydn,hou2024open}. The BEV-based methods transform the multi-view image features into a dense BEV grid representation and utilize it for the 3D object detection. However, dense BEV grid construction can substantially increase the computational complexity and GPU memory footprint of such methods. To address this, sparse-query-based methods introduce a set of sparse and learnable 3D object queries that directly attend to the multi-view image features, thereby eliminating the need for explicit BEV grid representation. Consequently, the computational complexity and GPU memory footprint are largely reduced, while achieving better performance compared to the dense BEV-based methods \cite{streampetr}.

Since sparse-query-based methods directly attend to the multi-view image features, the quality of multi-view image features are important for their performance \cite{toc3D}. As a result, large-scale pre-trained vision transformer (ViT)-based foundation models \cite{fang2024eva,SAM}, which demonstrate strong representation learning capabilities across many vision tasks, have been widely adopted as image encoders particularly in recent sparse-query-based methods \cite{sparse4dv2,streampetr,raydn}. By employing these image encoders \cite{fang2024eva,SAM}, sparse-query-based methods such as \texttt{Sparse4Dv2} \cite{sparse4dv2}, \texttt{StreamPETR} \cite{streampetr}, and \texttt{RayDN} \cite{raydn} achieve state-of-the-art (SOTA) results on the public 3D object detection benchmarks such as NuScenes \cite{caesar2020nuscenes} \cite{toc3D}. However, the high computational complexity of these ViT-based image encoders becomes the primary bottleneck in terms of inference speed, accounting for the majority of the total inference time in the sparse-query-based methods \cite{toc3D}. To address this bottleneck, the current SOTA method for efficient multi-view ViT-based 3D object detection, \texttt{ToC3D} \cite{toc3D} proposes an ego-motion-based image feature token selection module to accelerate ViT-based image encoders. The ego-motion-based image feature token selection module employs learnable motion queries, which are utilized to perform task-relevant token selection. Although it can be seamlessly integrated within any multi-view 3D object detection method, it employs fixed layer-individual token selection ratios in the ViT-based image encoder at both training and inference time, resulting in a suboptimal efficiency vs. performance trade-off. Here, efficiency is characterized by giga floating point operations (GFLOPs) and inference latency (ms), while performance is evaluated using standard 3D object detection metrics such as mean average precision (mAP) and NuScenes detection score (NDS) \cite{caesar2020nuscenes}. Motivated by this limitation, in this work, we focus on dynamic token selection to further push the efficiency vs. performance trade-off.

Beyond the fixed layer-individual number of selected tokens, \texttt{ToC3D} \cite{toc3D} also requires a full fine-tuning of the pre-trained multi-view 3D object detection model after integrating its ego-motion-based image feature token selection module within the ViT-based image encoder. Such a full fine-tuning is undesirable, as the original pre-trained weights of multi-view 3D object detection model are overwritten. Furthermore, full fine-tuning results in a large number of fine-tuned parameters, which significantly increases the training overhead, thereby reducing the training efficiency \cite{hu2022lora,lei2025rethinking}.

In this article, we make three contributions. First, we propose a \textit{dynamic layer-wise} token selection approach combined with an image token compensator adapted for the multi-view 3D object detection methods, which improves inference efficiency by reducing the computational complexity of the employed ViT-based image encoders, without sacrificing on the 3D object detection performance. Second, we propose a novel \textit{parameter-efficient fine-tuning} strategy, where we only train the proposed dynamic token selection modules, while freezing all other components of the multi-view 3D object detection method, thereby reducing the number of fine-tuned parameters from more than $300$ million (M) to only $1.6$M. Third, we show that our proposed approach decreases the GFLOPs down to $45\%$, inference latency down to $75\%$, while still slightly improving both mAP and NDS compared to \texttt{ToC3D} across three multi-view 3D object detection approaches on the large-scale NuScenes dataset \cite{caesar2020nuscenes}.

The rest of the article is structured as follows: Section \ref{sec:related_works} outlines all related works. Section \ref{sec:methods} provides a high-level overview of the basic multi-view 3D object detection pipeline, followed by the detailed description of our proposed methods. Section \ref{sec:experiments} describes the experimental setup, and evaluation metrics, including mAP, NDS, GFLOPs, and latency. Section \ref{sec:results} shows the experimental results. We conclude in Section \ref{sec:conclusion}.

\section{Related Works}
\label{sec:related_works}

This section provides an overview of multi-view 3D object detection, followed by a discussion on token selection methods for vision transformer-based image encoders, and parameter-efficient fine-tuning for multi-view 3D object detection.

\subsection{Multi-View 3D Object Detection}
Multi-view 3D object detection is a fundamental task in autonomous driving, which employs multi-view images from calibrated cameras to detect and localize objects in the surrounding environment of the ego-vehicle. Recent methods \cite{liftspatshoot,bevformerv1, bevformerv2,BEVdet,BEVDepth,Solofusion} compute a unified \textit{dense birds-eye-view (BEV) grid representation} from the multi-view image features, which is utilized for 3D object detection. \texttt{LSS} \cite{liftspatshoot} constructs BEV representations by predicting per-pixel depth distributions of the image features, while \texttt{BEVDet} \cite{BEVdet} introduces specialized data augmentation strategies and a scaled non-maximum suppression mechanism to improve \texttt{LSS} \cite{liftspatshoot} performance. \texttt{BEVDepth} \cite{BEVDepth} and \texttt{BEVStereo} \cite{li2023bevstereo} employ additional depth supervision in \texttt{LSS} to enhance its performance. \texttt{SOLOFusion} \cite{Solofusion} introduces short and long-term temporal fusion strategies in \texttt{BEVStereo} \cite{li2023bevstereo} and achieves state-of-the-art (SOTA) performance. Despite the strong performance of dense BEV-based methods, they have higher computational complexity and GPU memory footprint due to the requirement of dense BEV grid construction. To address this problem, sparse-query–based multi-view 3D object detection methods \cite{detr3d,petr,streampetr,mbev,sparse4d,sparsebev,raydn,hou2024open} have recently emerged, which employ a sparse set of learnable 3D object queries that directly interact with the multi-view image features to reduce both computational complexity and GPU memory footprint, while achieving better performance than dense BEV-based methods. \texttt{StreamPETR} \cite{streampetr} proposes a memory queue for storing sparse 3D object queries, enabling real-time feature propagation over time. \texttt{Sparse4Dv2} \cite{sparse4dv2} extends \texttt{DETR3D} \cite{detr3d} in the temporal domain by aggregating information across multiple timesteps through sparse 3D object queries. \texttt{RayDN} \cite{raydn} proposes a small set of learnable ray-casted queries to denoise 3D object queries. 

\textit{Sparse-query-based methods} \cite{detr3d,petr,streampetr,mbev,sparse4d,sparsebev,raydn,hou2024open} employ a sparse set of learnable 3D object queries in the 3D object decoder, which directly attend to the multi-view image features without constructing an explicit dense BEV grid representation. Consequently, the quality of the multi-view image features are important for the overall detection performance \cite{toc3D}. This motivates the use of the large-scale pre-trained vision transformer (ViT)-based foundation models \cite{SAM,fang2024eva} for multi-view image feature extraction. When combined with such image encoders, \texttt{StreamPETR} \cite{streampetr}, \texttt{RayDN} \cite{raydn}, and \texttt{Sparse4Dv2} \cite{sparse4dv2} achieve SOTA performance on the NuScenes benchmark \cite{caesar2020nuscenes} \cite{toc3D}. However, ViT-based image encoders have high computational complexity \cite{dosovitskiy2020image}, dominating the total runtime \cite{toc3D}. Therefore, in this work, we aim at efficiency improvement of the ViT-based image encoders by accelerating the multi-view 3D object detection methods, while maintaining competitive detection performance.

\subsection{Token Selection Methods for Vision Transformer-Based Image Encoders for Multi-View 3D Object Detection}

Vision transformers (ViTs) have demonstrated remarkable success across a wide range of computer vision tasks owing to their self attention mechanism, which enables them to model long-range dependencies and capture global relationships within the images. However, the self attention mechanism scales quadratically with the number of image feature tokens leading to high computational complexity. To alleviate this, current SOTA multi-view 3D object detection methods \cite{streampetr,raydn,sparse4d,sparse4dv2} constrain the self attention mechanism to local windows by adopting shifted window-based self attention \cite{liu2021swin} within the ViT-based image encoder. Nonetheless, the image encoder still operates on the full image feature token set, which contains image tokens that are irrelevant for the multi-view 3D object detection \cite{toc3D}.

In 2D computer vision tasks such as image classification \cite{deng2009imagenet}, semantic segmentation \cite{nazirjd,houben2022inspect}, and other related tasks, several image feature token selection methods \cite{dynamicvit,evo,tome,vidtldr,lei2025rethinking} have been proposed to reduce the computational burden of ViTs by discarding or merging less informative tokens while preserving downstream task performance. For the multi-view 3D object detection task, \texttt{tgGBC} \cite{tgbc} and \texttt{ToC3D} \cite{toc3D} have introduced image feature token strategies to accelerate the multi-view 3D object detection methods.~\texttt{tgGBC} \cite{tgbc} proposes a key-pruning-based image feature token selection strategy to accelerate 3D object decoders. However, it achieves limited gains in reducing the total computational complexity, since modern 3D object decoders incur significantly lower computational complexity than ViT-based image encoders \cite{toc3D}. To reduce the computational complexity of the ViT-based image encoders, \texttt{ToC3D} \cite{toc3D} proposes an ego-motion-based token selection module to select the relevant image feature tokens within the image encoders. Although it reduces total computational complexity, it introduces additional overhead by employing learnable motion queries for the image feature token selection. Also, it employs fixed layer-individual token selection ratios, which results in limited efficiency gains.

In this work, following \texttt{ToC3D} \cite{toc3D}, we aim at reducing the computational complexity of the image encoders through image feature token selection, but we eliminate the use of motion queries. Inspired by \cite{lei2025rethinking}, we propose a dynamic image feature token selection approach along with an image token compensator adapted for multi-view 3D object detection task.

\subsection{Parameter-Efficient Fine-Tuning for Multi-View 3D Object Detection}

Parameter-efficient fine-tuning (PEFT) techniques \cite{hu2022lora,zhao2024dynamic,lei2025rethinking} for 2D computer vision tasks aim at reducing training computational cost by fine-tuning only a small subset of large ViT-based pre-trained model parameters with lightweight modules or low-rank updates, while keeping the rest of the network frozen. The fine-tuned parameters typically contain multilayer perceptron-based or lightweight convolutional neural network-based decoders, together with the proposed modules by \cite{hu2022lora,zhao2024dynamic,lei2025rethinking}. However, such PEFT approaches are not directly applicable to multi-view 3D object detection, as modern multi-view 3D object detection methods \cite{streampetr,raydn,sparse4d,sparse4dv2,bevformerv1} employ complex decoders with multi-head self attention and cross attention mechanisms, making fine-tuning computationally expensive and less parameter-efficient.

For the multi-view 3D object detection task, recent work \texttt{ToC3D} \cite{toc3D} introduces a full fine-tuning strategy, where it fine-tunes the large pre-trained ViT-based image encoder, feature pyramid network, and the 3D object decoder of a multi-view 3D object detection method. However, such a full fine-tuning strategy is not parameter-efficient, as it significantly increases the training time computational complexity \cite{zhao2024dynamic,lei2025rethinking}. Therefore, in this work, we propose a novel PEFT strategy for multi-view 3D object detection methods by freezing the weights of the large pre-trained ViT-based image encoder, feature pyramid network, \textit{and} the 3D object decoder, while training only the proposed image feature token selection and compensator modules. Our approach significantly reduces the number of fine-tuned parameters, thereby lowering the GPU memory footprint during fine-tuning, while preserving the original trained weights of the respective multi-view 3D object detection method.

\section{Methods}
\label{sec:methods}
In this section, we first provide an overview of a basic multi-view 3D object detection pipeline, followed by baseline token selection for the ViT-based image encoder. Afterwards, we introduce our novel dynamic layer-wise token selection method, followed by our proposed parameter-efficient fine-tuning for the multi-view 3D object detection methods.

\subsection{Basic Multi-View 3D Object Detection Pipeline}
\label{sec:multiview_3d_obj_det_pipeline}

\begin{figure}[t!]
  \hspace{-2.7em} 
  \resizebox{1.1\linewidth}{!}{ \usetikzlibrary{positioning,shapes,arrows,arrows.meta,fit,backgrounds,calc}
 
 \definecolor{encoder_decoder}{RGB}{213, 232, 212}
 \definecolor{conv}{RGB}{255, 255, 255}
 \definecolor{bottleneck_color}{RGB}{253,211,177}
 \definecolor{pp_color}{RGB}{240, 240, 120}
 \definecolor{rectangle}{RGB}{204, 255, 153}

 \tikzstyle{arrow} = [-Triangle, line width=1pt]
 \tikzstyle{label} = [text width= 0.2cm, align=center]
 \tikzstyle{waypoint}=[fill,circle,minimum size=4.2pt,inner sep=0pt]
 
 \tikzstyle{encoder} = [trapezium,
    trapezium angle=55,
    trapezium stretches=true,
    minimum width=2.6cm,
    minimum height=1.12cm,
    trapezium right angle=80,
    trapezium left angle=80,
    line width=1pt,
    shape border rotate=270,
    text centered,
    draw=black]
    
\tikzstyle{decoder} = [trapezium,
trapezium angle=55,
trapezium stretches=true,
minimum width=2.6cm,
minimum height=1.12cm,
trapezium right angle=80,
trapezium left angle=80,
line width=1pt,
shape border rotate=90,
text centered,
draw=black]      
\tikzstyle{data} = [
    draw, 
    cylinder, 
    line width=1pt, 
    shape border rotate=90,
    cylinder uses custom fill, 
    aspect=0.17,
    cylinder body fill=white,
    cylinder end fill=white,
    minimum height=15pt,
    minimum width=30pt,
    outer sep=0pt, 
    inner sep=4pt,
    align=center
]

\tikzstyle{process} = [rectangle,minimum width=5cm, minimum height=1cm, text centered, draw=black, text=black,line width = 1pt, fill=conv]


\begin{tikzpicture} [node distance = 1.5cm,font=\fontsize{18}{20}\selectfont]

\node (true_bg) [draw, rectangle, minimum width=20cm, minimum height=4cm,  line width=1pt,draw=none, fill=none,yshift=0.cm] at (0,0) {};

\node (multi_view2) [data,xshift = 2.02cm,yshift=0.3cm, text=white,font=\fontsize{16.75}{18.75}\selectfont] at (true_bg.west)    {\shortstack{$\text{Multi-view}$ \\ [0.17em] $\text{Images}$}};

\node (multi_view) [data,font=\fontsize{16.75}{18.75}\selectfont]  at ([xshift=1.85cm,yshift=-0.6cm] true_bg.west)  {\shortstack{$\text{Multi-view}$ \\ [0.17em] $\text{Images}$}};

\node (input_way1)[waypoint] [xshift=0.6cm,yshift=-0.16cm, right of=multi_view2]  {}; 
\node (input_way2)[waypoint] [xshift=0.12cm,yshift=-1.35cm, above of=input_way1]  {}; 
\node (input_way3)[waypoint] [xshift=0.12cm,yshift=-1.35cm, above of=input_way2]  {};

\node (3d_object_detection) [draw, rectangle, minimum width=10.2cm, minimum height=4.0cm, line width=1pt, draw=none, fill=none, anchor=north,xshift=0cm, yshift=2.0cm] at (true_bg.center)  {};
\fill[color=rectangle, opacity=0.6, line width=1.5pt] ([yshift=0cm,xshift=0cm]3d_object_detection.north west) rectangle ([xshift=0cm,yshift=0cm]3d_object_detection.south east);
 \node[font=\Large,  label, text width=4cm, above of=3d_object_detection, align=center ,xshift=-2.1cm,yshift=0.15cm,font=\fontsize{16.75}{18.75}\selectfont ] {$\textbf{3D Object Detection Method}$};

\node (encoder2) [encoder,xshift=0.95cm, above=2.5cm of 3d_object_detection.south west, anchor=north west, fill=encoder_decoder, font=\fontsize{19.5}{21.5}\selectfont] {$\mathbf{E}$};

\node (encoder) [encoder,xshift=0.15cm, above=2cm of 3d_object_detection.south west, anchor=north west, fill=encoder_decoder, font=\fontsize{19.5}{21.5}\selectfont] {$\mathbf{E}$};

\node (encoder_way1)[waypoint] [xshift=-0.5cm,yshift=0.25cm, right of=encoder2]  {}; 
\node (encoder_way2)[waypoint] [xshift=0.12cm,yshift=-1.35cm, above of=encoder_way1]  {}; 
\node (encoder_way3)[waypoint] [xshift=0.12cm,yshift=-1.35cm, above of=encoder_way2]  {}; 

\node (fpn2) [process, xshift=3.77cm, yshift=0.96cm, right of =encoder,  minimum width=1.8cm, minimum height=0.9cm,fill=bottleneck_color,font=\fontsize{17.5}{19.5}\selectfont]  {\shortstack{$\mathbf{FPN}$  }};

\node (fpn) [process, xshift=2.1cm, yshift=0.0cm, right of =encoder,  minimum width=1.8cm, minimum height=0.9cm,fill=bottleneck_color,font=\fontsize{17.25}{19.25}\selectfont]  {\shortstack{$\mathbf{FPN}$  }};

\node (decoder_way1)[waypoint] [xshift=1.5cm,yshift=0.73cm, right of=fpn]  {}; 
\node (decoder_way2)[waypoint] [xshift=0.12cm,yshift=-1.35cm, above of=decoder_way1]  {}; 
\node (decoder_way3)[waypoint] [xshift=0.12cm,yshift=-1.35cm, above of=decoder_way2]  {};

\node (decoder) [decoder , right of=fpn2, xshift=2cm,yshift=-0.55cm, fill=encoder_decoder,font=\fontsize{19.5}{21.5}\selectfont] { \shortstack{$\mathbf{D}$}};
\node (nmsfree) [process, xshift=1.7cm, yshift=0.0cm, right of =decoder,  minimum width=2.7cm, minimum height=0.9cm,fill=pp_color,font=\fontsize{16.75}{18.75}\selectfont]  {\shortstack{\\ [0.05em]  $\text{Post}$$\text{-}$\\$\text{Processing}$  \\ [0.0em] }};
\draw[arrow] (multi_view) to  node[midway, above, align=center,xshift=0.06cm,yshift=-0.03cm,font=\fontsize{20}{22}\selectfont] {$\mathbf{x}_{1,t}$}   (encoder);
\draw[arrow] ([yshift=0.52cm]multi_view2.east) to  node[midway, above, align=center,xshift=0.0cm,yshift=-0.03cm,font=\fontsize{20}{22}\selectfont] {$\mathbf{x}_{v,t}$}   ([yshift=0.93cm]encoder2.west);
\draw[arrow] (encoder) to  node[midway, above, align=center,xshift=0.4cm,yshift=-0.03cm,font=\fontsize{20}{22}\selectfont] {$\mathbf{z}_{1,t}$}   (fpn);
\draw[arrow] ([yshift=0.7cm]encoder2.east) to  node[midway, above, align=center,xshift=0.cm,yshift=-0.03cm,font=\fontsize{20}{22}\selectfont] {$\mathbf{z}_{v,t}$}   ([yshift=0.24cm]fpn2.west);
\draw[arrow] (fpn.east) to  node[midway, above, align=center,xshift=0.4cm,yshift=-0.03cm,font=\fontsize{20}{22}\selectfont] {$\mathbf{r}_{1,t}$}   ([yshift=-0.41cm]decoder.west);
\draw[arrow] ([yshift=0.21cm]fpn2.east) to  node[midway, above, align=center,xshift=0.0cm,yshift=-0.03cm,font=\fontsize{20}{22}\selectfont] {$\mathbf{r}_{v,t}$}   ([yshift=0.76cm]decoder.west);
\draw[arrow] (decoder) to  node[midway, above, align=center,xshift=0cm,yshift=0.1cm,font=\fontsize{20}{22}\selectfont] {$\mathbf{y}_{t}$}   (nmsfree);
\draw[arrow] (nmsfree) -- ++(2.4,0) node[midway, xshift=-0.05cm, yshift=0.4cm,] {$\mathbf{u}_{t}$};

\end{tikzpicture}

}
    \caption{\textbf{High-level overview} of the basic multi-view $(v=1,2,\dots)$ 3D object detection pipeline used across multi-view 3D object detection methods, including \texttt{StreamPETR} \cite{streampetr}, \texttt{Sparse4Dv2} \cite{sparse4dv2}, and \texttt{RayDN} \cite{raydn} for training and inference. Here, $\mathbf{E}$ and $\mathbf{D}$ are the image encoder and task decoder, respectively. $\mathbf{FPN}$ represents a feature pyramid network. Details of the image encoder $\mathbf{E}$ are depicted in Fig.\ \ref{fig:wmhsa_comparison}. 
    }
    \label{fig:high_level_overview}
\end{figure}
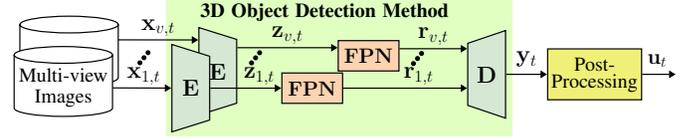

In Fig.\ \ref{fig:high_level_overview}, we illustrate a high-level overview of the basic multi-view 3D object detection pipeline used across multi-view 3D object detection methods, including \texttt{StreamPETR} \cite{streampetr}, \texttt{Sparse4Dv2} \cite{sparse4dv2}, and \texttt{RayDN} \cite{raydn} for both training and inference time. It consists of several components, including a view-individual image encoder $\mathbf{E}$, a view-individual feature pyramid network $\mathbf{FPN}$, a joint 3D object decoder $\mathbf{D}$, and a post-processing module. The image encoder $\mathbf{E}$ with parameters $\bm{\theta}^{\mathrm{E}}$ produces bottleneck features $\mathbf{z}_{v,t} = \mathbf{E}(\mathbf{x}_{v,t}; \bm{\theta}^{\mathrm{E}}) \in \mathbb{R}^{F \times \frac{H}{k} \times \frac{W}{k}}$ with camera view index $v \in \mathcal{V}$, camera view index set $\mathcal{V} = \{1,\dots,V\}$, number of camera views $V$, time index $t \in \mathcal{T}$ from index set $\mathcal{T} = \{1,\dots,T\}$, total number of frames $T$, kernels $F$, height $H$, width $W$, and image downsampling factor $k \in \mathbb{N}$. They are passed through a feature pyramid network $\mathbf{FPN}$ with parameters $\bm{\theta}^{\mathrm{FPN}}$ to compute multi-level pyramid features $\bf{r}_{\mathit{v,t}}=\mathbf{FPN}\big(\bf{z}_{\mathit{v,t}};\bm{\theta}^{\mathrm{FPN}}\big)$, where $\mathbf{r}_{v,t} = (\mathbf{r}_{v,t}^{(p)})_{p \in \mathcal{P}}$, with $\mathbf{r}_{v,t}^{(p)} \in \mathbb{R}^{F \times H_{p} \times W_{p}}$, pyramid index $p\in \mathcal{P}$ from index set $\mathcal{P} = \{1,\dots,P\}$, and the number of pyramid feature levels $P$. Here, $\mathbf{x}_{v,t} = (\mathbf{x}_{v,t,i}) \in \mathbb{I}^{\raisebox{-0.6ex}{$\scriptstyle C\times H\times W$}}$ is a normalized multi-view image of $C=3$ color channels, with pixel $\mathbf{x}_{v,t,i} \in \mathbb{I}^{C}$, pixel index $i \in \mathcal{I}$ from index set $\mathcal{I} = \{1,\dots,H \cdot W\}$, and $\mathbb{I} = [0,1]$. Weights for both $\mathbf{E}$ and $\mathbf{FPN}$ are shared among all $V$ camera views. The 3D object decoder $\mathbf{D}$ with parameters $\bm{\theta}^{\mathrm{D}}$ receives the multi-level pyramid features $\mathbf{r}_{t}=(\mathbf{r}_{v,t})_{v \in \mathcal{V}}$ and generates 3D bounding box classes $\mathbf{y}_{t}^{\mathrm{cls}}\in\mathbb{I}^{O \times S}$ and bounding box predictions $\mathbf{y}_{t}^{\mathrm{bbox}}\in\mathbb{R}^{O \times B}$ with the number of object queries $O$, bounding box elements $B$, class index $s \in \mathcal{S}$ from index set $\mathcal{S} = \{ 1,2,\dots,S \}$, and number of classes $S$, both jointly denoted as $\mathbf{y}_{t}=\big(\mathbf{y}_{t}^{\mathrm{cls}},\mathbf{y}_{t}^{\mathrm{bbox}}\big)=\mathbf{D}(\bf{r}_{\mathit{t}};\bm{\theta}^{\mathrm{D}})$. In the post-processing step, confidence-based filtering \cite{wang2022detr3d} is applied on $\mathbf{y}_{t}^{\mathrm{cls}}$ and $\mathbf{y}_{t}^{\mathrm{bbox}}$, resulting in $\mathbf{u}_{t}=\big(\mathbf{u}_{t}^{\mathrm{cls}},\mathbf{u}_{t}^{\mathrm{bbox}}\big)$. Note that in this article, following the current state-of-the-art method \texttt{ToC3D} \cite{toc3D} for efficient multi-view ViT-based 3D object detection, we employ ViT-based image encoders $\mathbf{E}$ and aim to reduce their computational complexity, which can be employed with any multi-view 3D object detection method.

\begin{figure}[t!]
  \hspace{-4.8em}
  \resizebox{1.19\linewidth}{!}{\input{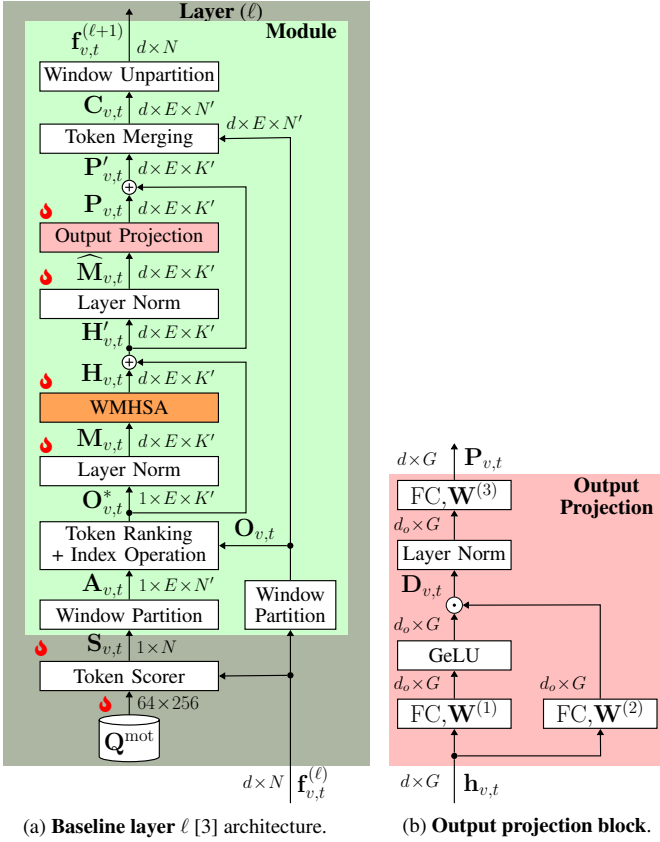}}
    \caption{(a) \textbf{Layer architecture} of the image encoder $\mathbf{E}$ utilized in Fig.\ \ref{fig:high_level_overview} for the baseline \cite{toc3D}. (b) Details the output projection block \cite{fang2024eva}. Blocks marked with \FireTok~are updated during fine-tuning, blocks with \IceTok~remain frozen.
    } 
    \label{fig:wmhsa_comparison}
\end{figure}

\subsection{Baseline Token Selection for ViT-Based Image Encoders}
\label{sec:baseline_layer_architecture}

This subsection lays foundations by describing the layer architecture of the ViT-based image encoder $\mathbf{E}$ in the current state-of-the-art (SOTA) method \texttt{ToC3D} \cite{toc3D}.

Fig.\ \ref{fig:wmhsa_comparison}(a) shows the network architecture of the baseline layer as used by the current SOTA method \texttt{ToC3D} \cite{toc3D} for efficient multi-view ViT-based 3D object detection. It employs a set of $64$ learnable motion queries $\mathbf{Q}^{\mathrm{mot}} \in \mathbb{R}^{64 \times 256}$ with internal dimension $256$ to capture the motion information of the objects using the ego-motion. The motion queries $\mathbf{Q}^{\mathrm{mot}}$ interact with the input image feature tokens  $\mathbf{f}_{v,t}^{(\ell)} \in \mathbb{R}^{d \times N}$ for image feature token selection, with layer index $\ell \in \mathcal{L}$, layer index set $\mathcal{L} = \{ 1,\dots,L\}$, number of layers $L$, internal dimension $d\in \mathbb{N}$, and number of image feature tokens $N\!=\!\frac{H}{k} \cdot \frac{W}{k} \in \mathbb{N}$. For $\ell=1$, the input image feature tokens represent the patch embedding layer's output\cite{dosovitskiy2020image}. The token scorer \cite{toc3D} receives learnable motion queries $\mathbf{Q}^{\mathrm{mot}}$ and $\mathbf{f}^{(\ell)}_{v,t}$ to compute saliency scores $\mathbf{S}_{v,t} \in \mathbb{I}^{1 \times N}$  with $\mathbb{I}\!=\![0,1]$ of each image feature token. Both, $\mathbf{S}_{v,t}$ and $\mathbf{f}_{v,t}^{(\ell)}$, are passed to a transformer module (module, cf. Fig.\ \ref{fig:wmhsa_comparison}(a)) \cite{liu2021swin}, where they are separately partitioned into windows through a reshape operation, yielding $\mathbf{A}_{v,t} = (A_{v,t,m}) \in \mathbb{I}^{1 \times E \times N^{\prime}}$ and $\mathbf{O}_{v,t}=(\mathbf{O}_{v,t,m}) \in \mathbb{R}^{d \times E \times N^{\prime}}$, respectively, with window size $f \in \mathbb{N}$, image feature tokens per window $N^{\prime}\!=\!f^{2}\!\in\!\mathbb{N}$, number of windows $E\!=\!\frac{H_p}{f} \cdot \frac{W_p}{f } \! \in \! \mathbb{N}$, zero-padded per-view image feature tokens height $H_p=h_p+\frac{H}{k} \in \mathbb{N}$, and width $W_p=w_p+\frac{W}{k} \in \mathbb{N}$. Here, $h_p,w_p \in \mathbb{N}$ represent the padding along the height and width dimensions, employed to compute equal-sized partitions of $\mathbf{f}^{(\ell)}_{v,t}$. Further, $\mathbf{O}_{v,t,m} \in \mathbb{R}^{d}$ denotes the window-partitioned image feature token, with respective index $m \in \mathcal{M}$ from index set $\mathcal{M}= \{ 1,\cdots, E \cdot N^{\prime}\}$. The window-partitioned saliency scores $\mathbf{A}_{v,t}$ are ranked corresponding to their saliency scores \cite{toc3D,dynamicvit} to obtain the ranked window-partitioned saliency scores $\mathbf{A}^{\prime}_{v,t} = ({A}^{\prime}_{v,t,c}) \in \mathbb{I}^{1\times E\times N^\prime}$, where $c \in \mathcal{C}$ denotes the respective index from index set $\mathcal{C}= \{ 1,\cdots, E \cdot N^{\prime}\}$. The highest ranked image feature tokens $\mathbf{O}^{\ast}_{v,t}$ are selected from $\mathbf{O}_{v,t}$ through an index operation \cite{dynamicvit} defined as $\mathcal{F}:\mathbb{R}^{d \times E \times N^{\prime}} \times \mathbb{I}^{1 \times E \times N^{\prime}} \rightarrow \mathbb{R}^{d \times E \times K^{\prime}}$ and $\mathbf{O}^{\ast}_{v,t} = \mathcal{F}(\mathbf{O}_{v,t}|\mathbf{A}_{v,t}^{\prime}) = \{ \mathbf{O}_{v,t,\pi(c)} |~c\leq K\} \in \mathbb{R}^{d \times E \times K^{\prime}}$. Here, $\pi : \mathcal{C}\rightarrow\mathcal{M}$ represents a permutation induced by the token ranking, which maps each ranked index $c \in \mathcal{C}$ to its corresponding index $m \in \mathcal{M}$ such that $A^{\prime}_{v,t,c} = A_{v,t,m=\pi(c)}$. Further, $K=E \cdot K^{\prime} \in \mathbb{N}$ denotes the number of selected image feature tokens across all windows, and $K^{\prime}\in\mathbb{N}$ is the number of selected image feature tokens per window such that $K^{\prime}\!<\! N^{\prime}$. Note that in the baseline layer \cite{toc3D}, $K$ is not input-dependent and determined by a predefined layer-specific token selection ratio that is kept constant during both training and inference. The selected image feature tokens $\mathbf{O}^{\ast}_{v,t}$ are normalized using a layer norm to obtain $\mathbf{M}_{v,t} \in \mathbb{R}^{d \times E \times K^{\prime}}$, and then processed by a window-based multi-head self attention (WMHSA) block with $I=K^{\prime}$ for the baseline to produce a refined representation $\mathbf{H}_{v,t} \in \mathbb{R}^{d \times E \times I}$, which is used to generate the residual output as $\mathbf{H}^{\prime}_{v,t} = \mathbf{O}_{v,t}^{\ast} + \mathbf{H}_{v,t} \in \mathbb{R}^{d \times E \times K^{\prime}}$. Note that the WMHSA block now operates on $K$ image feature tokens, which reduces its computational complexity. However, the WMHSA block accounts for only a small fraction of the total computational complexity of the image encoder layer, resulting in limited overall efficiency gains. The WMHSA block details are described in Appendix \ref{app:wmhsa}. The residual output $\mathbf{H}^{\prime}_{v,t}$ is further passed through a layer norm, producing $\widehat{\mathbf{M}}_{v,t}$, which is passed to the output projection block with the inputs $\mathbf{h}_{v,t}=\widehat{\mathbf{M}}_{v,t}$ and $G\!=K$, described in the following.  
\begin{table}[t!]
  \caption{\textbf{Computational complexity breakdown of a naive transformer module} in the \texttt{EVA-02-L} \cite{fang2024eva} encoder without any image feature token selection, measured with an input resolution of $320\!\times\!800$ on $\mathcal{D}^{\mathrm{val}}_{\mathrm{NS}}$.
  }
  \label{table:complexity_window_attention}
  \centering
    \setlength{\tabcolsep}{4pt} 
    \begin{tabular} {lrrrr}
        \toprule
               \textbf{Block} & \textbf{FLOPs (G)}  & \textbf{$\#$params (M)} \\
            \midrule
                WMHSA & $6.9$ & $4.2$ \\
                Output Projection & $16.8$ & $8.4$ \\
            \midrule
                Total & $23.7$ & $12.6$ \\
        \bottomrule
    \end{tabular}
\end{table}

\noindent\textbf{Output Projection Block:} Fig.\ \ref{fig:wmhsa_comparison}(b) depicts the output projection block. The input $\mathbf{h}_{v,t} \in  \mathbb{R}^{d \times G}$ is projected as 
\begin{equation}
\mathbf{D}_{v,t}=\mathrm{GeLU}\big(\mathbf{W}^{(1)} \mathbf{h}_{v,t}\big) \odot \big(\mathbf{W}^{(2)} \mathbf{h}_{v,t} \big) , 
\end{equation}
to compute the projected image feature tokens $\mathbf{D}_{v,t} \in \mathbb{R}^{d_o \times G}$, with learnable FC layers weights $\mathbf{W}^{(1)}, \mathbf{W}^{(2)} \in \mathbb{R}^{d_{o} \times d}$, hidden dimension $d_{o} \in \mathbb{N}$, element-wise multiplication $\odot$, and the $\mathrm{GeLU}$ activation function. The projected image feature tokens $\mathbf{D}_{v,t}$ are passed through a layer norm, followed by an output FC projection layer with weights $\mathbf{W}^{(3)} \in \mathbb{R}^{d \times d_{o}}$ to compute the output projection block output $\mathbf{P}_{v,t} \in \mathbb{R}^{d \times G}$, which is added to $\mathbf{H}^{\prime}_{v,t}$ producing the residual output $\mathbf{P}_{v,t}^{\prime} \in \mathbb{R}^{d\times G}$. Note that in the baseline layer, the output projection block operates on the number of selected image feature tokens across all windows $K$, and due to zero-padding, $K$ becomes larger than the number of image feature tokens $N$ \cite{fang2024eva}. Further, the computational complexity of the output projection block scales linearly with the number of input image feature tokens \cite{zhao2024dynamic,lei2025rethinking}. Therefore, \textit{the computational complexity of the output projection block is significantly increased}.

\noindent\textbf{Token Merging:} Back to Fig.\ \ref{fig:wmhsa_comparison}(a), in token merging, $\mathbf{P}^{\prime}_{v,t}$ is merged with $\mathbf{O}_{v,t}$ to produce the merged image feature tokens $\mathbf{C}_{v,t} \in \mathbb{R}^{d \times E \times N^{\prime}}$. Through token merging, the spatial positions of $\mathbf{P}^{\prime}_{v,t}$ are restored to their corresponding spatial locations in $\mathbf{O}_{v,t}$. The merged image feature tokens $\mathbf{C}_{v,t}$ are reshaped and the zero-padding is removed to obtain the baseline layer \cite{toc3D} output $\mathbf{f}^{(\ell + 1)}_{v,t}$.

\begin{figure}[t!]
  \hspace{-4.5em}
  \resizebox{1.19\linewidth}{!}{\input{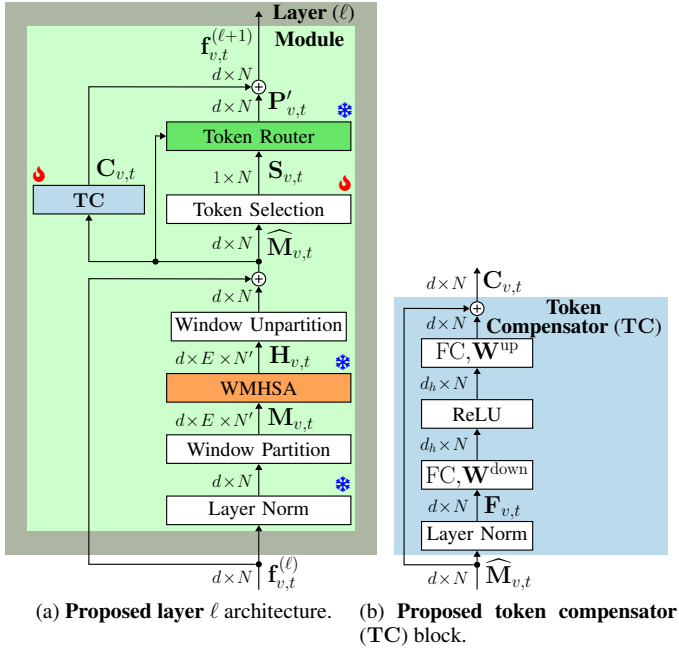}}
    \caption{(a) \textbf{Proposed layer architecture} of the image encoder $\mathbf{E}$ utilized in Fig.\ \ref{fig:high_level_overview}. (b) \textbf{Proposed token compensator} $(\mathbf{TC})$ block. Blocks marked with \FireTok~are updated during fine-tuning, blocks with \IceTok~remain frozen. 
    } 
    \label{fig:proposed_layer_tc}
\end{figure}
\subsection{Proposed Dynamic Layer-wise Token Selection}
\label{sec:proposed_layer_architecture}

Now, we motivate the design of our proposed layer for ViT-based image encoders and then describe it in detail.

\noindent\textbf{Motivation}: The current SOTA method \texttt{ToC3D} \cite{toc3D} for efficient multi-view ViT-based 3D object detection employs learnable motion queries $\mathbf{Q}^{\mathrm{mot}}$ to select the subset of important image feature tokens $K$ in $\mathbf{f}^{(\ell)}_{v,t}$. Although it reduces the computational complexity of the image encoder $\mathbf{E}$ layer, it suffers from two key limitations: (1) The number of selected tokens $K$ is not input-dependent and remains fixed for each layer during both training and inference, limiting its adaptability to diverse scenes. (2) The output projection block (cf. Fig.\ \ref{fig:wmhsa_comparison}(b)) operates on $K$ image feature tokens, where $K>N$, resulting in limited efficiency gains, when employed with the multi-view 3D object detection methods. As shown in Table \ref{table:complexity_window_attention}, the WMHSA block in a naive \texttt{EVA-02-L} \cite{fang2024eva} transformer module, which does not perform any image feature token selection, exhibits significantly lower computational complexity than the output projection block. It accounts for only $29 \%$  of the FLOPs ($6.9~\mathrm{G}$) and $33 \%$ of the parameters ($4.2~\mathrm{M}$), compared to the $71\%$ of FLOPs ($16.8~\mathrm{G}$) and $67\%$ of parameters ($8.4~\mathrm{M}$) of the output projection block. Therefore, inspired by \cite{lei2025rethinking}, \textit{we propose a layer architecture with dynamic image token selection adapted for the multi-view 3D object detection task. Our proposed layer is fully input-dependent and enables dynamic image feature token selection in each layer of $\mathbf{E}$ for the output projection block, allowing it to operate on a smaller set of image feature tokens and thereby achieving substantial efficiency gains without any noticeable loss of performance. Further, our design also simplifies the baseline layer \cite{toc3D} described in Section \ref{sec:baseline_layer_architecture} by omitting the use of $\mathbf{Q}^{\mathrm{mot}}$ for the image feature token selection}.

 \begin{figure}[t!]
\centering
  \hspace{-4.0em}
  \resizebox{1.13\linewidth}{!}{\input{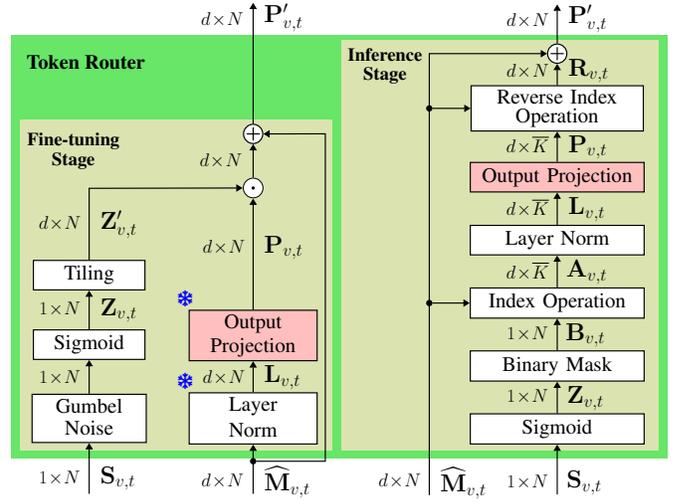}}
    \caption{\textbf{Our token router block} used in Fig.\ \ref{fig:proposed_layer_tc}(a), adopted from \cite{zhao2024dynamic,lei2025rethinking}. The output projection block (details in Fig.\ \ref{fig:wmhsa_comparison}(b)) remains frozen \IceTok~during the fine-tuning stage.}
    \label{fig:token_router}
\end{figure}

\noindent\textbf{Proposed Layer Architecture}: Fig.\ \ref{fig:proposed_layer_tc}(a) illustrates our proposed layer architecture. The image feature tokens $\mathbf{f}^{(\ell)}_{v,t} \in \mathbb{R}^{d \times N}$ are directly passed to the transformer module (module), which applies layer norm and partitions it to windows by applying zero-padding and a reshape operation to compute $\mathbf{M}_{v,t} \in \mathbb{R}^{d \times E\times N^{\prime}}$. The window-partitioned image feature tokens $\mathbf{M}_{v,t}$ are passed to the WMHSA block with $I\!=\!N^{\prime}$ to generate its refined representation $\mathbf{H}_{v,t} \in \mathbb{R}^{d \times E \times N^\prime}$. The details of the WMHSA block are described in Appendix \ref{app:wmhsa}. Further, the refined representation $\mathbf{H}_{v,t}$ is reshaped and the zero-padding is removed before it is added to $\mathbf{f}_{v,t}^{(\ell)}$ to produce $\widehat{\mathbf{M}}_{v,t} \in \mathbb{R}^{d \times N}$, followed 
by a image feature token selection block consisting of a single FC-based layer to compute the saliency scores $\mathbf{S}_{v,t} \in \mathbb{I}^{1 \times N}$ \cite{zhao2024dynamic} of each image feature token. The saliency scores $\mathbf{S}_{v,t}$ together with $\mathbf{\widehat{M}}_{v,t}$ are passed to the token router block. 

\noindent\textbf{Proposed Token Router}: Fig.\ \ref{fig:token_router} details the token router block \cite{lei2025rethinking,zhao2024dynamic}. It contains distinct pathways for fine-tuning and inference stages. During the \textit{fine-tuning stage}, a layer normalization is applied to $\widehat{\mathbf{M}}_{v,t}$ to obtain $\mathbf{L}_{v,t}$, which is sent to the output projection block that projects $\mathbf{L}_{v,t}$ to $\mathbf{P}_{v,t} \in \mathbb{R}^{d \times N}$ with $\mathbf{h}_{v,t}=\mathbf{L}_{v,t}$ as its input and number of per-view image feature tokens $G=N$. In parallel, the saliency scores $\mathbf{S}_{v,t}$ are passed through a Gumbel noise approximation to enable the backpropagation \cite{zhao2024dynamic,lei2025rethinking}, followed by a sigmoid activation function to normalize $\mathbf{S}_{v,t}$ as $\mathbf{Z}_{v,t}=\mathbf{\bm{\sigma}}(\mathbf{S}_{v,t})\in \mathbb{I}^{1 \times N}$ with $\mathbb{I}=[0,1]$. The normalized saliency scores $\mathbf{Z}_{v,t} \in \mathbb{I}^{1 \times N}$ are subject to tiling along the internal feature dimension $d$ according to $\mathbf{Z}^{\prime}_{v,t}=\mathbf{1} \otimes \mathbf{Z}_{v,t} \in \mathbb{I}^{d \times N}$, whereby the Kronecker product $\otimes$ produces $d$ copies of $\mathbf{Z}_{v,t}$ on top of each other. This is used to generate the output of the token router block as $\mathbf{P}^{\prime}_{v,t} = \mathbf{Z}^{\prime}_{v,t} \odot \mathbf{P}_{v,t} ~+ \widehat{\mathbf{M}}_{v,t} \in \mathbb{R}^{d \times N}$. Here, $\odot$ denotes element-wise multiplication. 
 
 During the \textit{inference stage} (Fig.\ \ref{fig:token_router}), the Gumbel noise approximation is removed and the saliency scores $\mathbf{S}_{v,t}$ are directly passed through the sigmoid function to generate the normalized saliency scores $\mathbf{Z}_{v,t}=\mathbf{\bm{\sigma}}(\mathbf{S}_{v,t})\in \mathbb{I}^{1 \times N}$. The normalized saliency scores $\mathbf{Z}_{v,t}=(Z_{v,t,j})$ are thresholded to generate a binary mask $\mathbf{B}_{v,t} = (B_{v,t,j}) \in \mathcal{B}^{1 \times N}$ with $B_{v,t,j} =  Z_{v,t,j}>\theta$, $\mathcal{B}=\{0,1\}$, normalized saliency score $Z_{v,t,j} \in \mathbb{I}$, with index $j \in \mathcal{J}$ from index set $\mathcal{J} = \{ 1,\dots, N \}$. The binary mask $\mathbf{B}_{v,t}$ is used to select the important image feature tokens $\mathbf{A}_{v,t}$ from $\widehat{\mathbf{M}}_{v,t}=(\widehat{\mathbf{M}}_{v,t,j})$ through an index operation \cite{zhao2024dynamic} defined as $\mathcal{F}\!:\! \mathbb{R}^{d \times N} \times \mathcal{B}^{1 \times N} \to \mathbb{R}^{d \times \overline{K}}$ and $\mathbf{A}_{v,t} = \mathcal{F}(\widehat{\mathbf{M}}_{v,t} | \mathbf{B}_{v,t}) \in \mathbb{R}^{d \times \overline{K}}$, with $\mathcal{F}(\widehat{\mathbf{M}}_{v,t}|\mathbf{B}_{v,t})= \{ \widehat{\mathbf{M}}_{v,t,j}| {B}_{v,t,j}=1 \}$, and number of image feature tokens $\overline{K}$. The selected image feature tokens $\mathbf{A}_{v,t}$ are passed through a layer norm to obtain $\mathbf{L}_{v,t} \in \mathbb{R}^{d \times \overline{K}}$, which is passed through an output projection block with the input $\mathbf{h}_{v,t}=\mathbf{L}_{v,t}$, and number of selected image feature tokens $G=\overline{K}$. It produces the projected output $\mathbf{P}_{v,t}=(\mathbf{P}_{v,t,k}) \in \mathbb{R}^{d\times \overline{K}}$ with projected image feature token $\mathbf{P}_{v,t,k} \in \mathbb{R}^{d}$, index $k \in \mathcal{K}$, from index set $\mathcal{K} = \{ 1, \dots , \overline{K} \} $. Using $\mathbf{B}_{v,t}$ and $\overline{K}$, the indices of selected image feature tokens in $\mathcal{J}$ are computed as $\mathcal{J}^{\ast} =\{ j ~|~B_{v,t,j}=1\} \subseteq \mathcal{J}$ with selected image feature token index $j^{\ast} \in \mathcal{J}^{\ast}$, and their total number $|\mathcal{J}^{\ast}|=\overline{K}$. The detailed network architecture of the output projection block matches the baseline and is shown in Figure \ref{fig:wmhsa_comparison}(b).  

\noindent\textbf{Reverse Index Operation}: The projected image feature tokens $\mathbf{P}_{v,t}$ are passed through a reverse index operation \cite{zhao2024dynamic} defined as $\mathcal{V}: \mathbb{R}^{d \times \overline{K}} \rightarrow \mathbb{R}^{d \times N}$ and $\mathbf{R}_{v,t} =  \mathcal{V}(\mathbf{P}_{v,t}) = (\mathbf{R}_{v,t,j}) \in \mathbb{R}^{d \times N}$  with 
     \begin{equation}
    \mathbf{R}_{v,t,j} = \begin{cases}
     \mathbf{P}_{v,t,\pi(j)}, & \text{if } j \in \mathcal{J}^{\ast},\\[2mm]
\mathbf{0}=\{0\}^{d}, & \text{otherwise},
 \end{cases}  ,
    \end{equation}
to restore the spatial structure of $\widehat{\mathbf{M}}_{v,t}$. Here, the permutation $\pi: \mathcal{J}^{\ast} \rightarrow \mathcal{K}$ denotes the mapping between each selected original image feature token index in $\mathcal{J}^{\ast}$ with its corresponding index in $\mathcal{K}$, enabling the selected tokens to be restored to their original spatial locations.

In contrast to the baseline layer described in Section \ref{sec:baseline_layer_architecture}, where the output projection block operates on $K$ image feature tokens with $K>N$, our proposed layer applies it only to the selected image feature tokens $\overline{K}$. Consequently, the computational complexity of our output projection block is significantly reduced, resulting in substantial overall efficiency gains, without any noticeable loss of performance. Note that $\overline{K}$ is fully input-dependent and dynamic for each layer in $\mathbf{E}$, as it is determined by a learnable binary mask $\mathbf{B}_{v,t}$ generated from $\mathbf{Z}_{v,t}$. Further, in parallel to the image feature token selection layer and output projection block, we propose to employ a token compensator block, which is described in the following. 

\noindent\textbf{Proposed Token Compensator ($\mathbf{TC}$)}:  Fig.\ \ref{fig:proposed_layer_tc}(b) details the proposed token compensator $\mathbf{TC}$ block. The goal of the $\mathbf{TC}$ block is to compensate for any loss of contextual information during the image feature token selection process and refine the selected image feature tokens for the multi-view 3D object detection task by learning to reconstruct $\widehat{\mathbf{M}}_{t}$. Unlike prior work \cite{lei2025rethinking} that adopts a convolutional neural network-based block for token compensation suitable for 2D computer vision tasks, we propose a simple \textit{two-layered FC block} that restores the contextual information after image feature token selection and achieves even higher performance for the multi-view 3D object detection task. The token compensator $\mathbf{TC}$ with parameters $\bm{\theta}^{\mathrm{TC}}$ applies layer norm to $\widehat{\mathbf{M}}_{v,t}$ to produce $\mathbf{F}_{v,t} \in \mathbb{R}^{d \times N}$, which is projected through two separate FC projection layers 
 \begin{equation}
     \mathbf{F}^{\prime}_{v,t} =  \mathbf{W}^{\mathrm{up}} \cdot \mathrm{ReLU}\big(\mathbf{W}^{\mathrm{down}} \mathbf{F}_{v,t}\big) 
 \end{equation}
\noindent to compute $\mathbf{F}_{v,t}^{\prime} \in \mathbb{R}^{d \times N}$, with learnable FC layers weights $\mathbf{W}^{\mathrm{down}} \in \mathbb{R}^{d_{h} \times d}$ and $\mathbf{W}^{\mathrm{up}} \in \mathbb{R}^{d \times d_{h}}$, hidden dimension $d_{h}$, and a $\mathrm{ReLU}$ activation function. The projected output $\mathbf{F}_{v,t}^{\prime}$ is added to $\widehat{\mathbf{M}}_{v,t}$ to generate the $\mathbf{TC}$ output $\mathbf{C}_{v,t} \in \mathbb{R}^{d \times N}$.

The aggregated information $\mathbf{C}_{v,t}$ is added to $\mathbf{P}^{\prime}_{v,t}$ to compute the proposed layer output $\mathbf{f}^{(\ell + 1)}_{v,t}$. Following prior works \cite{zhao2024dynamic,lei2025rethinking}, we employ an average image feature token activation rate $r \in [0,1]$ to constrain the number of activated image feature tokens within the proposed layer.

\subsection{Proposed Parameter-Efficient Fine-Tuning (PEFT)}
 \label{sec:peft}
 The current SOTA method \texttt{ToC3D} \cite{toc3D} for efficient multi-view ViT-based 3D object detection employs learnable motion queries $\mathbf{Q}^{\mathrm{mot}}$ for image feature token selection, which requires full fine-tuning of the pretrained multi-view 3D object detection model. This approach has two key limitations. (1) The pretrained weights of the multi-view 3D object detection models are overwritten. (2) Full fine-tuning of the multi-view 3D object detection significantly increases the training overhead due to the large number of trainable parameters. Therefore, to address these limitations, \textit{we propose a novel parameter-efficient fine-tuning (PEFT) strategy for multi-view 3D object detection methods, which fine-tunes only the proposed image feature token selection and token compensation blocks (TC) in each layer of encoder $\mathbf{E}$, Fig.\ \ref{fig:proposed_layer_tc}(a) in comparison to Fig.\ \ref{fig:wmhsa_comparison}(a), while keeping the rest of the model weights frozen.} Furthermore, our proposed PEFT strategy truly enables a plug-and-play design, as removing these blocks exactly restores the original pretrained multi-view 3D object detection model.

\section{Experimental Setup}
\label{sec:experiments}
In this section, we introduce the dataset. Afterwards, we provide a detailed overview of our training and evaluation settings, followed by the introduction of the standard 3D object detection and efficiency metrics.

\subsection{Dataset}

Following our primary baselines, \texttt{ToC3D} \cite{toc3D}, and \texttt{tgGBC} \cite{tgbc}, we conduct all our experiments on the well-established and widely adopted NuScenes dataset for multi-view 3D object detection in autonomous driving. The official NuScenes training split ($\mathcal{D}_{\mathrm{NS}}^{\mathrm{train}}$) contains $700$ scenes with $168,100$ images, while the official validation split ($\mathcal{D}_{\mathrm{NS}}^{\mathrm{val}}$) consists of $150$ scenes with $36,000$ images. Each scene is captured using six high-resolution calibrated cameras at $2$Hz frame rate, providing a full $360^{\circ}$ field of view around the ego vehicle. All scenes are densely annotated with 3D bounding boxes for 10 common object classes and additional attributes, supporting tasks such as multi-view 3D object detection, tracking, and motion prediction. Since the official NuScenes benchmark server for the test set is no longer functional, we follow recent publication \texttt{tgGBC} \cite{tgbc} and report all results on $\mathcal{D}_{\mathrm{NS}}^{\mathrm{val}}$.

\subsection{Experimental Design and Training}

We employ \texttt{SAM-B} \cite{SAM} and \texttt{EVA-02-L} \cite{fang2024eva} as our ViT-based image encoders $\mathbf{E}$ to evaluate our proposed method vs.\ important prior works \cite{toc3D,tgbc}. \texttt{SAM-B} represents a lightweight encoder, while \texttt{EVA-02-L} corresponds to a larger and heavyweight encoder. \textit{Note that recent multi-view 3D object detection approaches \cite{streampetr,bevformerv2,hou2024open,raydn,sparse4dv2,li2024bevnext} predominantly adopt \texttt{EVA-02-L} \cite{fang2024eva} as image encoder, which has become the most widely used ViT-based image encoder}. Both, \texttt{ToC3D} \cite{toc3D} and our approach, only modify $\mathbf{E}$ in the multi-view 3D object detection methods by integrating either the baseline layer (Section \ref{sec:baseline_layer_architecture}) or the proposed layer (Section \ref{sec:proposed_layer_architecture}), respectively. In contrast, \texttt{tgGBC} \cite{tgbc} modifies the 3D object decoder $\mathbf{D}$, while keeping $\mathbf{E}$ and the feature pyramid network $\mathbf{FPN}$ unchanged. Following \texttt{ToC3D} \cite{toc3D}, we adopt \texttt{StreamPETR} \cite{streampetr} as our primary method for multi-view 3D object detection, but we also propose to validate the generalizeability and effectiveness of our proposed layer on two other methods, namely \texttt{Sparse4Dv2} \cite{sparse4dv2} and \texttt{RayDN} \cite{raydn}. For \texttt{ToC3D} \cite{toc3D}, we employ the token selection ratios of $0.7,0.5,0.5$ in $\mathbf{E}$ for both training and evaluation, as it provides the strongest performance among the reported variants. Similarly, we integrate \texttt{tgGBC} \cite{tgbc} into the \texttt{StreamPETR} \cite{streampetr} method and perform retraining, as this configuration achieves the highest performance. Further, in all our experiments, we set the number of camera views to $V=6$, the number of bounding box elements to $B=9$, the hidden dimension $d_o = \lfloor2.66~\cdot d \rfloor$ \cite{fang2024eva} in the output projection block (cf.\ Fig.\ \ref{fig:wmhsa_comparison}(b)), and the binary mask threshold $\theta=0.5$ \cite{zhao2024dynamic}. The detailed training and evaluation settings for \texttt{StreamPETR}, \texttt{Sparse4Dv2}, and \texttt{RayDN} on the NuScenes \cite{caesar2020nuscenes} dataset are described in the following.     

\noindent\textbf{\texttt{StreamPETR}}: On \texttt{StreamPETR} \cite{streampetr}, we perform the experiments with both \texttt{SAM-B} and \texttt{EVA-02-L} image encoders. For \texttt{SAM-B} \cite{SAM}, we utilize an image resolution of $320 \times 800$ and evaluate our method vs. the \texttt{tgGBC} \cite{tgbc} baseline. Similarly, for \texttt{EVA-02-L} \cite{fang2024eva}, we employ the image resolutions of $320 \times 800$ and $800 \times 1600$, and evaluate our proposed approach vs. \texttt{ToC3D} \cite{toc3D}, and \texttt{tgGBC} \cite{tgbc} baselines. We follow the same settings suggested by the two baselines \cite{toc3D},\cite{tgbc}.  

\noindent\textbf{\texttt{Sparse4Dv2} and \texttt{RayDN}}: For \texttt{Sparse4Dv2} \cite{streampetr}, we evaluate our proposed approach on the image resolution of $320 \times 800$ with an \texttt{EVA-02-L} \cite{fang2024eva} image encoder as suggested by the \texttt{ToC3D} baseline. We apply the same configuration also to the \texttt{RayDN} \cite{raydn} to maintain consistency across both methods.

To ensure a fair comparison, all baseline results are reproduced using the respective open-source implementations provided by the authors, including trainings according to our employed methods \cite{streampetr,raydn,sparse4dv2}. All trainings are conducted on an \texttt{NVIDIA-H100} GPU, while the evaluations are performed on a lightweight \texttt{NVIDIA-GV100} GPU to simulate real-world deployment scenarios. 

\subsection{Metrics}
\label{section:metrics}
To evaluate 3D object detection performance, we employ the following standard detection and efficiency metrics. 

\noindent\textbf{Mean Average Precision:} The predicted 3D bounding boxes are evaluation using the mean average precision \cite{caesar2020nuscenes}
 \begin{equation}
    \mathrm{mAP} = \frac{1}{S\cdot U} \sum_{s \in \mathcal{S}}\sum_{u \in \mathcal{U}} AP_{u,s} ~ ~,
    \label{eq:map}
\end{equation}where $u \in \mathcal{U}=\{1, \cdots,U\}$ is the 3D bounding box threshold index, $\mathcal{U}$ is the 3D bounding box threshold index set, $U$ being the total number of such thresholds, and $AP_{u,s}$ is the average precision \cite{caesar2020nuscenes} for 3D bounding box threshold $u$ and class $s \in \mathcal{S}=\{1,\dots,S\}$.

\noindent\textbf{NuScenes Detection Score:} The NuScenes detection score (NDS) is the primary evaluation metric of the NuScenes benchmark. It combines mAP with $W=5$ true positive (TP) error metrics \cite{caesar2020nuscenes}, including average translation error (ATE), average scale error (ASE), average orientation error (AOE), average velocity error (AVE), and average attribute error (AAE) to jointly evaluate the localization, scale, orientation, velocity, and attribute accuracy of the predicted 3D object bounding boxes. For each TP error metric, the mean is computed over all classes as
\begin{equation}
      \mathrm{mTP}_{w} = \frac{1}{S} \sum_{s \in \mathcal{S}} TP_{w,s} ~ ~, 
    \label{eq:mtp_w}
\end{equation}
with TP error metric index $w \in \mathcal{W}= \{ 1, \dots, W \}$, and TP error metric $TP_{w,s}$ for class $s$. Using (\ref{eq:map}) and (\ref{eq:mtp_w}), the NDS is defined as \begin{equation}
      \mathrm{NDS} = \frac{1}{2}\mathrm{mAP} + \frac{1}{10} \sum_{w \in \mathcal{W}} (1-\mathrm{min(1,mTP}_{w})) ~.
\end{equation}

\noindent\textbf{GFLOPs:} For multi-view 3D object detection methods, giga floating point operations (GFLOPs) measure the total number of floating point arithmetic operations required to perform a single forward pass, given $\mathbf{x}_{t} = (\mathbf{x}_{v,t})_{v \in \mathcal{V}}$. It serves as a standard metric for reporting model computational complexity \cite{nazirjd,zhao2024dynamic,lei2025rethinking}.

\noindent\textbf{Average Rank:} To provide a comparison across all standard evaluation and efficiency metrics, we additionally report average (avg.) rank \cite{Urgent1,Urgent2}. The average rank is computed by ranking all methods in a table segment for each metric independently (higher is better for mAP/NDS, lower is better for TP error metrics, model size, GFLOPs, and latency), and row-wise averaging the ranks across all metrics. Lower average rank indicates better overall performance.

\noindent\textbf{Latency:} For multi-view 3D object detection methods, latency is a critical metric because it determines how quickly the method can react to the dynamic changes in the environment. Following the current SOTA method \texttt{ToC3D} \cite{toc3D} for efficient multi-view ViT-based 3D object detection, we define the latency of the multi-view 3D object detection as

\begin{equation}
      \tau = \tau^{\mathrm{E}} + \tau^{\mathrm{FPN}} + \tau^{\mathrm{D}}
\label{eq:latency_results}
\end{equation}

\noindent where $\tau^{\mathrm{E}}$, $\tau^{\mathrm{FPN}}$, $\tau^{\mathrm{D}}$, and $\tau$ denote the wall-clock times required to compute bottleneck features $\mathbf{z}_{t}=(\mathbf{z}_{t,v})_{v \in \mathcal{V}}$ from $\mathbf{x}_{t}=(\mathbf{x}_{t,v})_{v \in \mathcal{V}}$, multi-level pyramid features $\mathbf{r}_{t}$ from $\mathbf{z}_{t}$, final 3D bounding box predictions $\mathbf{u}_{t}$ from $\mathbf{r}_{t}$, and final 3D bounding box predictions $\mathbf{u}_{t}$ from $\mathbf{x}_{t}$, respectively. Latency is measured on an \texttt{NVIDIA-GV100} GPU. Note that the frames per second (fps) can be simply computed by $\mathrm{fps}=\frac{1}{\tau}$.

\begin{table}[t!]
  \caption{\textbf{Ablation study to select the internal hidden dimension $d_{h}$} of the proposed token compensator utilized in Fig.\ \ref{fig:wmhsa_comparison}(b) on $\mathcal{D}^{\mathrm{val}}_{\mathrm{NS}}$ with an \texttt{EVA-02-L} \cite{fang2024eva} encoder. The encoder employs $r=0.5$ and is integrated into \texttt{StreamPETR} \cite{streampetr}. The encoder-only ($\tau^{\mathrm{E}}$) and complete pipeline latencies ($\tau$) are reported. Best results bold, second-best underlined.}
  \label{table:as_select_d_TC}
  \centering
    \setlength{\tabcolsep}{4pt} 
    \begin{tabular} {rrrrrr}
        \toprule
               $d_{h}$  
              & \multirow{2}{*}{\makecell{\textbf{mAP} \textuparrow \\ (\%)}}  
              & \multirow{2}{*}{\makecell{\textbf{NDS} \textuparrow \\ (\%)}}  
              & \multirow{2}{*}{\makecell{\textbf{GFLOPs} \textdownarrow \\} }
              & \multicolumn{2}{c}{\makecell{\textbf{Latency} (ms) \textdownarrow \\  }}  \\
              \cmidrule(lr){5-6}
            & & & & $\tau^{\mathrm{E}}$ & $\tau$  \\
            \midrule
                 $16$ & $50.1$ & $58.9$ & $\mathbf{5,754}$ & $\mathbf{444}$ & $\mathbf{485}$  \\
                 $32$ & $\mathbf{50.9}$  & $\mathbf{59.6}$ & $\underline{5,759}$ & $\underline{454}$ & $\underline{492}$  \\
                 $64$ & $50.8$ & $59.5$ & $5,860$ & $496$ & $537$  \\
        \bottomrule
    \end{tabular}
\end{table}

\begin{figure}[t!]
  \hspace{-1.35em}
  \resizebox{1.062\linewidth}{!}{\input{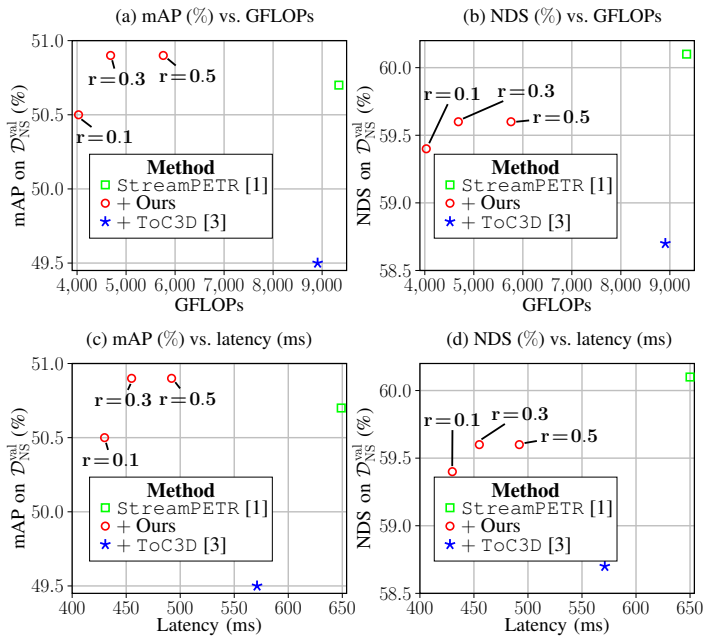}}
    \caption{\textbf{Comparison of mAP and NDS} in dependence on GFLOPs and latency (ms), measured with an input resolution of $320\times800$ for $\mathcal{D}^{\mathrm{val}}_{\mathrm{NS}}$ for our proposed approach vs. \texttt{ToC3D} \cite{toc3D}, both integrated into \texttt{StreamPETR} \cite{streampetr}. Symbol $r$ represents the average image feature token activation rate in the image encoder layer.}
    \label{fig:results_diagram}
\end{figure}

\section{Experimental Results and Discussion}
\label{sec:results}
\subsection{Ablation Study}

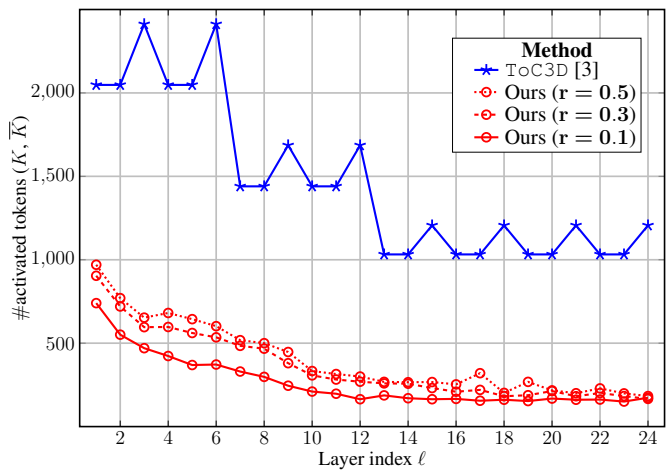
\begin{figure}[t!]
  \hspace{-1.4em}
  \resizebox{1.03\linewidth}{!}{

\definecolor{lightblue}{RGB}{166,206,227}   
\definecolor{sandybrown}{RGB}{253,191,111}  
\definecolor{lightgreen}{RGB}{178,223,138}  
\definecolor{lightsalmon}{RGB}{251,154,153} 


\pgfplotstableread[col sep=comma]{
layer,r01,r03,r05,ToC3D
1,740,904,970, 2048
2,551,720,770, 2048
3,470,596,653, 2412
4,423,597,681, 2048
5,369,561,644, 2048
6,372,535,602, 2412
7,330,484,518, 1440
8,298,467,500, 1440
9,245,380,447, 1686
10,210,307,333, 1440
11,197,282,315, 1440
12,164,269,300, 1686
13,187,259,268, 1032
14,170,257,266, 1032
15,164,231,267, 1206
16,166,209,253, 1032
17,155,220,320, 1032
18,161,182,202, 1206
19,153,188,268, 1032
20,168,213,217, 1032
21,161,182,200, 1206
22,162,201,229, 1032
23,150,183,199, 1032
24,173,166,183, 1206
}\datatable


\begin{tikzpicture}
\begin{axis}[
    name=activated_tokens, 
    width=16.18cm, height=12cm,
    x grid style={gray!50, line width=1.2pt},
    y grid style={gray!50, line width=1.2pt},
    axis line style={black, line width=1.15pt},
    ymin=0, ymax=2500,
    ymajorgrids,
    ytick distance=500,
    ytick={500,1000,1500,2000},
    ylabel style={yshift=8pt},
    xlabel style={yshift=1pt},
    grid=major, 
    grid style={gray!30},
    tick label style={color=black,font=\fontsize{14}{16}\selectfont},
    label style={font=\fontsize{14}{16}\selectfont},
    xlabel={Layer index $\ell$},
    ylabel={$\#$activated tokens ($K$, $\overline{K}$)},
    xmajorgrids,
    xmin=1, xmax=24,
    xtick distance=2,
    xtick style={color=black},
    enlarge x limits=0.03,
    legend cell align=left,
    legend style={
        draw=none, fill=none, font=\small,
        legend columns=4,
        /tikz/every even column/.append style={column sep=6mm}
    },
]

\addplot+[mark=o,line width=1.5pt,  red,  mark=o, mark size=3.1,  mark options={ line width=1.3, solid}] table[x=layer,y=r01] {\datatable};
\label{act_ours_r_0_1}

\addplot+[mark=o,dashed,line width=1.5pt,  red,  mark=o, mark size=3.1,  mark options={ line width=1.3, solid}] table[x=layer,y=r03] {\datatable};
\label{act_ours_r_0_3}

\addplot+[mark=o,line width=1.5pt, dotted, red,  mark=o, mark size=3.1,  mark options={ line width=1.3, solid}] table[x=layer,y=r05] {\datatable};
\label{act_ours_r_0_5}

\addplot+[mark=o,line width=1.5pt,  blue,  mark=star, mark size=4.2,  mark options={ line width=1.3, solid}] table[x=layer,y=ToC3D] {\datatable};

\label{act_toc3d_legend}






\end{axis}

\node[draw,fill=white,font=\fontsize{15.2}{17.2}\selectfont, anchor=north west] at ([xshift=2.cm,yshift=4.5cm]activated_tokens) {
\shortstack[l]{
\hspace{3.em}\textbf{Method} \\
\hspace{0.75em}\ref*{act_toc3d_legend} \texttt{ToC3D} \cite{toc3D}  \\
\hspace{0.75em}\ref*{act_ours_r_0_5} Ours ($\mathbf{r\boldsymbol{=}0.5}$)  \\
\hspace{0.75em}\ref*{act_ours_r_0_3} Ours ($\mathbf{r\boldsymbol{=}0.3}$)  \\
\hspace{0.75em}\ref*{act_ours_r_0_1} Ours ($\mathbf{r\boldsymbol{=}0.1}$) 
}
};

\end{tikzpicture}

}
    \caption{\textbf{Comparison of $\#$activated tokens at the output projection block} (Fig.\ \ref{fig:wmhsa_comparison}(b)) in each image encoder layer, measured with an input resolution of $320 \times 800$ on 300 randomly sampled instances from $\mathcal{D}^{\mathrm{val}}_{\mathrm{NS}}$ for our proposed approach vs.\ \texttt{ToC3D}\cite{toc3D}, both integrated into \texttt{StreamPETR} \cite{streampetr}.}
    \label{fig:activated_tokens}
\end{figure}

In Table \ref{table:as_select_d_TC}, we present an ablation study to select the internal hidden dimension $d_{h}$ of the proposed token compensator $\mathbf{TC}$ (Fig.\ \ref{fig:proposed_layer_tc}(b)) at an input resolution of $320\!\times\!800$ on $\mathcal{D}^{\mathrm{val}}_{\mathrm{NS}}$ with an \texttt{EVA-02-L} \cite{fang2024eva} encoder integrated into \texttt{StreamPETR} \cite{streampetr}. We ablate over $d_{h} \in \{16,32,64\}$ with a fixed image token activation rate $r\!=\!0.5$, applied in each transformer block of encoder \texttt{EVA-02-L}. As shown in Table \ref{table:as_select_d_TC}, the configuration with $d_{h}=32$ produces the best performance on mAP and NDS, with GFLOPs and latency being close-by to the $d_{h}=16$ case. Accordingly, for our proposed $\mathbf{TC}$ we select $d_{h}=32$. Interestingly, note that the selected value $d_{h}\!=\!32$ is applied consistently across \texttt{SAM-B} encoder and multiple multi-view 3D object detection methods, including \texttt{RayDN} and \texttt{Sparse4Dv2}. 

\subsection{Comparison to State-of-the-Art}

\begin{table*}[t!]
  \caption{\textbf{Comparison of mAP, NDS, and various 3D object detection error metrics} in dependence on GFLOPs and latency (ms), measured with two input resolutions on $\mathcal{D}^{\mathrm{val}}_{\mathrm{NS}}$ for various methods. In each table segment, best results bold, second-best underlined.}
  \centering
  \setlength{\tabcolsep}{3.1pt} 
\begin{tabular}{cllrrrrrrrrrrrr}
    \toprule
\multirow{2}{*}{\rotatebox{90}{\textbf{Enc.}}} 
& \multirow{2}{*}{\textbf{Methods}} 
& \multirow{2}{*}{\makecell{\textbf{Image} \\ \textbf{Resolution}}} 
& \multirow{2}{*}{\makecell{\textbf{mAP} \textuparrow \\ (\%) }} 
& \multirow{2}{*}{\makecell{\textbf{NDS} \textuparrow \\ (\%) }}  
& \multirow{2}{*}{\textbf{mATE} \textdownarrow} 
& \multirow{2}{*}{\textbf{mASE} \textdownarrow} 
& \multirow{2}{*}{\textbf{mAOE} \textdownarrow} 
& \multirow{2}{*}{\textbf{mAVE} \textdownarrow}  
& \multirow{2}{*}{\textbf{mAAE} \textdownarrow}
& \multirow{2}{*}{\makecell{\textbf{Model} \\ \textbf{Size} \textdownarrow \\ (M)}}
& \multirow{2}{*}{\textbf{GFLOPs} \textdownarrow} 
& \multicolumn{2}{c}{\makecell{\textbf{Latency} \textdownarrow}} 
& \multirow{2}{*}{\makecell{\textbf{Avg.} \\ \textbf{rank}~\textdownarrow}} \\
\cmidrule(lr){13-14}
& & & & & & & & & & & & $\tau^{\mathrm{E}}$ & $\tau$  \\
            \midrule
                 \multirow{3}{*}{\rotatebox{90}{\texttt{RN-50}}} & \texttt{BEVDet} \cite{BEVdet}   & $320\!\times\!800$ & $31.5$ & $39.2$ & $0.703$ & $0.269$ & $0.580$ & $\underline{0.864}$ & $\underline{0.243}$ & $\mathbf{48.1}$ &  $\mathbf{1,\!191}$ & $\mathbf{17}$ & $\mathbf{53}$ & $\underline{2.09}$  \\
               
                & \texttt{BEVDepth} \cite{BEVDepth}   & $320\!\times\!800$ & $\underline{34.6}$ & $\underline{40.2}$ & $\underline{0.639}$ & $\mathbf{0.267}$ & $\underline{0.556}$ & $0.984$ & $0.258$  & $\underline{53.6}$ & $\underline{1,\!349}$ & $\underline{37}$ & $\underline{79}$ & $\underline{2.09}$ \\
                
                & \texttt{SOLOFusion} \cite{Solofusion}   & $320\!\times\!800$ & $\mathbf{42.6}$ & $\mathbf{53.6}$ & $\mathbf{0.582}$ & $\underline{0.272}$ & $\mathbf{0.460}$ & $\mathbf{0.261}$ & $\mathbf{0.197}$ & $65.3$ & $1,\!451$ & $40$ & $567$ & $\mathbf{1.81}$  \\
              \midrule
              \midrule
               \multirow{4}{*}{\rotatebox{90}{\texttt{SAM-B} \cite{SAM}}} & \texttt{StreamPETR} \cite{streampetr}   & $320\!\times\!800$ & $\mathbf{46.3}$ & $\mathbf{55.5}$ & $\mathbf{0.615}$ & $\underline{0.266}$ & $0.417$ & $\underline{0.267}$ & $0.204$ & $\mathbf{99.7}$ & $3,\!546$ & $166$ & $205$ & $2.63$  \\

               & $+$ \texttt{tgGBC} \cite{tgbc}   & $320\!\times\!800$ & $43.4$ & $54.2$ & $0.638$ & $0.275$ & $\mathbf{0.388}$ & $\mathbf{0.244}$ & $0.215$ & $\mathbf{99.7}$ & $3,\!530$ & $159$ & $188$ & $2.81$ \\
               
               & $+$ Ours ($r=0.5$)   & $320\!\times\!800$ & $\underline{46.1}$ & $\underline{55.3}$ & $\underline{0.626}$ & $\mathbf{0.265}$ & $\underline{0.410}$ & $0.268$ & $\underline{0.203}$ & $\underline{100.2}$ & $2,\!679$ & $141$ & $179$ & $\mathbf{2.27}$ \\

                & $+$ Ours ($r=0.3$)   & $320\!\times\!800$ & $46.0$ & $55.1$ & $0.633$ & $\mathbf{0.265}$ & $0.422$ & $0.268$ & $\underline{0.203}$  & $\underline{100.2}$ & $\underline{2,\!449}$ & $\underline{129}$ & $\underline{173}$ & $\underline{2.45}$  \\

               & $+$ Ours ($r=0.1$) & $320\!\times\!800$ & $44.1$ & $53.9$ & $0.654$ & $\underline{0.266}$ & $0.424$ & $0.271$ & $\mathbf{0.200}$ & $\underline{100.2}$ & $\mathbf{2,\!295}$ & $\mathbf{123}$ & $\mathbf{160}$ & $2.81$ \\
        
              \midrule
              \midrule
               \multirow{12}{*}{\rotatebox{90}{\texttt{EVA-02-L} \cite{fang2024eva}}} & \texttt{StreamPETR} \cite{streampetr}   & $320\!\times\!800$ & $\underline{50.7}$ & $\mathbf{60.1}$ & $\mathbf{0.560}$ & $\mathbf{0.257}$ & $\mathbf{0.259}$ & $0.255$ & $\underline{0.196}$ & $\mathbf{316.6}$ & $9,\!342$ & $603$ & $650$ & $2.90$  \\
               
               & $+$ \texttt{tgGBC} \cite{tgbc} & $320\!\times\!800$ & $47.6$ & $57.8$ & $0.613$ & $0.260$ & $\underline{0.275}$ & $\underline{0.237}$ & $0.210$ & $\mathbf{316.6}$ & $9,\!325$ & $583$ & $607$ & $3.72$  \\
               
               & $+$ \texttt{ToC3D} \cite{toc3D} & $320\!\times\!800$ & $49.5$ & $58.7$ & $0.589$ & $0.260$ & $0.309$ & $0.253$ & $\underline{0.196}$ & $324.4$ & $8,\!908$ & $526$ & $571$ & $3.72$  \\

               & $+$ Ours ($r=0.5$)   & $320\!\times\!800$ & $\mathbf{50.9}$ & $\underline{59.6}$ & $\underline{0.585}$ & $\underline{0.258}$ & $0.304$ & $\underline{0.237}$ & $\underline{0.196}$ & $\underline{318.2}$ & $5,\!759$ & $454$ & $492$ & $\underline{2.27}$ \\

               & $+$ Ours ($r=0.3$)   & $320\!\times\!800$ & $\mathbf{50.9}$ & $\underline{59.6}$ & $\underline{0.585}$ & $\underline{0.258}$ & $0.306$ & $\mathbf{0.235}$ & $\mathbf{0.195}$ & $\underline{318.2}$ & $\underline{4,\!684}$ & $\underline{417}$ & $\underline{455}$ & $\mathbf{2.00}$  \\

               & $+$ Ours ($r=0.1$) & $320\!\times\!800$ & $50.5$ & $59.4$ & $0.593$ & $\mathbf{0.257}$ & $0.298$ & $0.241$ & $\mathbf{0.195}$ & $\underline{318.2}$ & $\mathbf{4,\!030}$ & $\mathbf{391}$ & $\mathbf{430}$ & $\mathbf{2.00}$  \\
             \cmidrule{2-15}

              & \texttt{StreamPETR} \cite{streampetr} & $800\!\times\!1600$ & $\mathbf{55.5}$ & $\mathbf{63.0}$ & $\underline{0.539}$ & $\underline{0.255}$ & $0.258$ & $0.231$ & $0.199$ & $\mathbf{316.6}$ & $45,\!044$ & $2,\!571$ & $2,\!724$ & $3.36$  \\

              & $+$ \texttt{tgGBC} \cite{tgbc} & $800\!\times\!1600$ & $54.2$ & $\underline{62.1}$ & $0.560$ & $0.256$ & $0.255$ & $\mathbf{0.223}$ & $0.204$ & $\mathbf{316.6}$ & $44,\!964$ & $2,\!491$ & $2,\!550$ & $3.54$   \\
               
              & $+$ \texttt{ToC3D} \cite{toc3D} & $800\!\times\!1600$ & $\underline{55.2}$ & $\mathbf{63.0}$ & $\mathbf{0.534}$ & $0.257$ & $\mathbf{0.241}$ & $0.231$ & $0.200$ & $324.4$ & $37,\!060$ & $2,\!020$ & $2,\!171$ & $2.90$  \\

              & $+$ Ours ($r=0.5$)  & $800\!\times\!1600$ & $\mathbf{55.5}$ & $\mathbf{63.0}$ & $\underline{0.539}$ & $\underline{0.255}$ & $0.256$ & $\underline{0.230}$ & $\underline{0.196}$ & $\underline{318.2}$ & $26,\!798$ & $1,\!954$ & $2,\!098$ & $\underline{2.27}$  \\

              & $+$ Ours ($r=0.3$)  & $800\!\times\!1600$ & $\mathbf{55.5}$ & $\mathbf{63.0}$ & $0.541$ & $\mathbf{0.254}$ & $0.258$ & $0.231$ & $0.197$ & $\underline{318.2}$  & $\underline{21,\!131}$ & $\underline{1,\!737}$ & $\underline{1,\!880}$ & $\underline{2.27}$  \\    
              
               & $+$  Ours ($r=0.1$)  & $800\!\times\!1600$ & $54.1$ & $\underline{62.1}$ & $0.556$ & $\mathbf{0.254}$ & $\underline{0.254}$ & $0.234$ & $\mathbf{0.194}$ & $\underline{318.2}$ & $\mathbf{18,\!512}$ & $\mathbf{1,\!652}$ & $\mathbf{1,\!789}$ & $\mathbf{2.09}$ \\  
              \midrule
        \bottomrule
    \end{tabular}
    \label{table:quantitative_results}
\end{table*}

In Fig.\ \ref{fig:results_diagram}, we illustrate the comparison of mAP and NDS in dependence on GFLOPs and latencies, measured with an input resolution of $320 \times 800$ on $\mathcal{D}^{\mathrm{val}}_{\mathrm{NS}}$ for our proposed method vs.\ \texttt{ToC3D} \cite{toc3D}, both employing the \texttt{EVA-02-L} \cite{fang2024eva} image encoder. Both, our proposed method and \texttt{ToC3D}, are integrated into the \texttt{StreamPETR} \cite{streampetr} method. Our method is reported for three distinct average image feature token activation rates $r$.
The first overall observation is that the mAP and NDS performance of all investigated methods is very similar when plotted over GFLOPs (upper two plots) or over latency (lower two plots). This means that higher computational complexity linearly translates into higher latency for the examined methods---which is not necessarily the case for any network topology. When comparing our approaches to the \texttt{StreamPETR} baseline \cite{streampetr}, the baseline is slightly better in NDS, and slightly worse w.r.t.\ mAP (for $r\in\{0.3, 0.5\}$), however, paying a high price in both GFLOPs and latency. \textit{Compared to the so-far SOTA \texttt{ToC3D} \cite{toc3D}, our approach is significantly better in both metrics mAP and NDS, while being significantly less complex and faster for all choices of $r$.}

In Fig.\ \ref{fig:activated_tokens}, we compare the number of activated tokens (\texttt{ToC3D}: $K$ vs.\ ours: $\overline{K}$) at the output projection block of each layer of the \texttt{EVA-02-L} \cite{fang2024eva} image encoder, measured with an input resolution of $320 \times 800$ on 300 randomly sampled instances from $\mathcal{D}^{\mathrm{val}}_{\mathrm{NS}}$. Both, our proposed method and \texttt{ToC3D} \cite{toc3D}, are integrated into the \texttt{StreamPETR} \cite{streampetr} method. For all methods, we observe a general trend of a decreasing number of activated tokens towards later layers of the encoder. Note that the number of activated tokens is closely related to the computational complexity of that particular layer. \textit{Comparing \texttt{ToC3D} token activations to ours, we observe that in early layers we activate about half of \texttt{ToC3D's} tokens ($\overline{K}\approx K/2$), while in late encoder layers, we activate only about $20\%$ of \texttt{ToC3D's} tokens ($\overline{K}\approx K/5$), thereby explaining the much lower computational complexity and latency of our method in Fig.\ \ref{fig:results_diagram}.} The choice of hyperparameter $r$ easily allows to control the number of activated tokens of our method in fine-tuning. 

Now we would like to investigate the value of using our proposed token compensator, shown in Fig.\ \ref{fig:proposed_layer_tc}(b). Table \ref{table:as_select_TC} reports a comparison of mAP and NDS in dependence on GFLOPs and latency measured with an input resolution of $320 \times 800$ on $\mathcal{D}^{\mathrm{val}}_{\mathrm{NS}}$ for the proposed token compensator $\mathbf{TC}$ (Fig.\ \ref{fig:proposed_layer_tc}(b)) vs.\ the one from \texttt{DAR-TR-PEFT} \cite{lei2025rethinking}. Both, our approach and \texttt{DAR-TR-PEFT}, utilize a \texttt{SAM-B} \cite{SAM} image encoder and are integrated into \texttt{StreamPETR} \cite{streampetr}. In Table \ref{table:as_select_TC}, we observe that the baseline token compensator from \texttt{DAR-TR-PEFT} is slightly less complex and faster to compute than ours. However, for all average image feature token selection rates $r$, our $\mathbf{TC}$ is consistently better than \texttt{DAR-TR-PEFT} \cite{lei2025rethinking} w.r.t.\ both mAP and NDS. This is particularly significant for $r=0.3$, where we clearly excel the baseline in mAP ($46.0\%$ vs.\ $43.6\%$) and in NDS ($55.1\%$ vs.\ $53.6\%$). Comparing across methods and $r$ values, we observe the exact same performance metrics (mAP = $46.0\%$, NDS = $55.1\%$) for our $\mathbf{TC}$ (at $r=0.3$) and the baseline (at $r=0.5$), while our token compensator is less complex and faster to execute.
\textit{Overall, the results demonstrate that the proposed $\mathbf{TC}$ consistently improves detection performance (mAP and NDS) at a close-by computational cost for fixed $r$, and enables a better efficiency vs.\ detection performance trade‑off by achieving the same performance at reduced computational cost across different values of $r$.}

\begin{table}[t!]
  \caption{\textbf{Comparison of mAP, NDS} in dependence on GFLOPs and latency (ms), on $\mathcal{D}^{\mathrm{val}}_{\mathrm{NS}}$ for our proposed token compensator ($\mathbf{TC}$, Fig.\ \ref{fig:proposed_layer_tc}(b)) vs.\ the token compensator by \texttt{DAR-TR-PEFT} \cite{lei2025rethinking}. Both, our approach and \texttt{DAR-TR-PEFT} \cite{lei2025rethinking} are integrated into \texttt{StreamPETR} \cite{streampetr} employing the \texttt{SAM-B} \cite{SAM} encoder. Best results bold. Latency in ms. }
  \centering
    \setlength{\tabcolsep}{4pt} 
    \begin{tabular} {llrrrrr}
        \toprule
              \textbf{Token Compensator}  
              &  $r$
              & \multirow{2}{*}{\makecell{\textbf{mAP} \textuparrow \\ (\%)}}  
              & \multirow{2}{*}{\makecell{\textbf{NDS} \textuparrow \\ (\%)}}  
               & \multirow{2}{*}{\makecell{\textbf{GFLOPs} \textdownarrow}}
              & \multicolumn{2}{c}{\makecell{\textbf{Latency} \textdownarrow \\ }} \\
              \cmidrule(lr){6-7}
              & & & & & $\tau^{\mathrm{E}}$ & $\tau$   \\
            \midrule
                \texttt{DAR-TR-PEFT} \cite{lei2025rethinking} & \multirow{2}{*}{$0.1$} & $43.0$ & $53.4$ & $\mathbf{2,\!291}$ & $\mathbf{121}$ &$\mathbf{158}$  \\
               Ours & & $\mathbf{44.1}$ & $\mathbf{53.9}$ &  $2,\!295$ & $123$ & $160$  \\
            \midrule
                 \texttt{DAR-TR-PEFT} \cite{lei2025rethinking} &  \multirow{2}{*}{$0.3$} & $43.6$ & $53.6$ & $\mathbf{2,\!447}$ & $\mathbf{128}$ & $\mathbf{168}$ \\
                 Ours &  & $\mathbf{46.0}$ & $\mathbf{55.1}$ & $2,\!449$  & $129$ & $173$  \\
            \midrule
                  \texttt{DAR-TR-PEFT} \cite{lei2025rethinking} & \multirow{2}{*}{$0.5$} & $46.0$ & $55.1$ &  $\mathbf{2,\!677}$ & $\mathbf{140}$ &$\mathbf{180}$ \\
                 Ours &  & $\mathbf{46.1}$ & $\mathbf{55.3}$ & $2,\!680$ & $141$ & $179$  \\
            \midrule
        \bottomrule
    \end{tabular}
    \label{table:as_select_TC}
\end{table}

\begin{table*}[t!]
  \caption{\textbf{Generalization study of our proposed method vs.\ \texttt{ToC3D} on two further 3D object detection methods}, \texttt{RayDN} \cite{raydn} and \texttt{Sparse4Dv2} \cite{sparse4dv2}. Evaluation metrics mAP, NDS, and various 3D object detection errors in dependence on GFLOPs and latencies (ms) are measured with an input resolution of $320\!\times\!800$ on $\mathcal{D}^{\mathrm{val}}_{\mathrm{NS}}$. In each table segment, best results bold, second-best underlined.}
  \centering
  \setlength{\tabcolsep}{4pt} 
    \begin{tabular} {lrrrrrrrrrrrr}
        \toprule 
 \multirow{2}{*}{\textbf{Methods}} 
& \multirow{2}{*}{\makecell{\textbf{mAP} \textuparrow \\ (\%) }} 
& \multirow{2}{*}{\makecell{\textbf{NDS} \textuparrow \\ (\%) }}  
& \multirow{2}{*}{\textbf{mATE} \textdownarrow} 
& \multirow{2}{*}{\textbf{mASE} \textdownarrow} 
& \multirow{2}{*}{\textbf{mAOE} \textdownarrow} 
& \multirow{2}{*}{\textbf{mAVE} \textdownarrow}  
& \multirow{2}{*}{\textbf{mAAE} \textdownarrow} 
& \multirow{2}{*}{\makecell{\textbf{Model} \\ \textbf{Size} \textdownarrow \\ (M)}}
& \multirow{2}{*}{\textbf{GFLOPs} \textdownarrow}  
& \multicolumn{2}{c}{\makecell{\textbf{Latency} \textdownarrow \\ }}
& \multirow{2}{*}{\makecell{\textbf{Avg.} \\ \textbf{rank} \textdownarrow }}\\ 
\cmidrule(lr){11-12}
& & & & & & & & & & $\tau^{\mathrm{E}}$ & $\tau$  \\ 
   
               \midrule
                \texttt{RayDN} \cite{raydn} &  $53.6$ & $\mathbf{62.3}$ & $0.540$ & $0.257$ & $\mathbf{0.250}$ & $\mathbf{0.201}$ & $0.204$ & $\mathbf{331.9}$ & $9,\!787$ & $615$ & $639$ & $3.09$  \\
                
                $+$ \texttt{ToC3D} \cite{toc3D} &  $50.9$ & $60.9$ & $0.523$ & $\mathbf{0.251}$ & $\underline{0.254}$ & $0.238$ & $\mathbf{0.190}$& $339.6$ & $9,\!359$ & $490$ & $510$ & $3.18$  \\
                
                $+$ Ours ($r=0.5$)   &  $\mathbf{54.0}$ & $\mathbf{62.3}$ & $\mathbf{0.519}$ & $\underline{0.252}$ & $0.275$ & $\underline{0.231}$ & $0.193$ & $\underline{333.5}$ & $6,\!396$ & $495$ & $518$ & $\mathbf{2.27}$ \\
                
                $+$ Ours ($r=0.3$)   &  $\underline{53.8}$ & $\underline{62.1}$ & $0.523$ & $\underline{0.252}$ & $0.274$ & $0.233$ & $\underline{0.192}$ & $\underline{333.5}$ & $\underline{5,\!484}$ & $\underline{480}$ & $\underline{503}$ & $\underline{2.45}$  \\
                
                $+$ Ours ($r=0.1$)   &  $53.7$ & $62.1$ & $\underline{0.522}$ & $\underline{0.252}$ & $0.275$ & $0.234$ & $0.193$ & $\underline{333.5}$ & $\mathbf{4,\!909}$ & $\mathbf{442}$ & $\mathbf{464}$ & $\mathbf{2.27}$ \\
             
              \midrule
              \midrule

                 \texttt{Sparse4Dv2} \cite{sparse4dv2} &  $\mathbf{53.8}$ & $\mathbf{61.3}$ & $\mathbf{0.529}$ & $\underline{0.259}$ & $\underline{0.292}$ & $0.287$ & $\underline{0.193}$ & $\mathbf{316.5}$ & $9,004$ & $246$ & $279$ & $\underline{2.54}$  \\
                
                 $+$ \texttt{ToC3D} \cite{toc3D} &  $50.7$ & $60.1$ & $0.559$ & $\mathbf{0.257}$ & $\mathbf{0.259}$ & $\mathbf{0.255}$ & $0.196$ & $324.1$ & $8,682$  & $221$ & $257$ & $3.36$ \\
                
                 $+$ Ours ($r=0.5$)   &  $\underline{53.5}$ & $\underline{61.0}$ & $0.533$ & $\underline{0.259}$ & $0.294$ & $0.292$ & $\mathbf{0.190}$ & $\underline{318.1}$ & $5,751$ & $220$ & $252$ & $2.63$ \\
                
                $+$ Ours ($r=0.3$)   &  $53.0$ & $60.8$ & $0.541$ & $\underline{0.259}$ & $0.288$ & $\underline{0.286}$ & $\underline{0.193}$ & $\underline{318.1}$ & $\underline{4,674}$ & $\underline{207}$ & $\underline{239}$ & $\mathbf{2.27}$ \\
                
                $+$ Ours ($r=0.1$) &  $52.5$ & $60.5$ & $0.546$ & $\underline{0.259}$ & $0.294$ & $0.289$ & $0.194$ & $\underline{318.1}$ & $\mathbf{3,879}$ & $\mathbf{198}$ & $\mathbf{230}$ & $\underline{2.54}$ \\
              \midrule
        \bottomrule
    \end{tabular}
    \label{table:generalization_study}
\end{table*}

In Table \ref{table:quantitative_results}, we show a detailed comparison of mAP, NDS, and the 3D object detection error metrics in dependence on model size, GFLOPs, and latencies, measured with input resolutions of $320\!\times\!800$ and $800\!\times\!1600$ on $\mathcal{D}^{\mathrm{val}}_{\mathrm{NS}}$ for our proposed method vs.\ \texttt{ToC3D} \cite{toc3D} and \texttt{tgGBC} \cite{tgbc}. We evaluate both our approach and \texttt{tgGBC} using image encoders with low and high computational complexity, namely \texttt{SAM-B} \cite{SAM} and \texttt{EVA-02-L} \cite{fang2024eva}, respectively. Further, \texttt{ToC3D} is evaluated with an \texttt{EVA-02-L} encoder. Our proposed method and the baselines \texttt{tgGBC} and \texttt{ToC3D} are integrated into \texttt{StreamPETR} \cite{streampetr}. We report and rank results separately in table segments depending on image resolution and encoders. For a fair comparison, we also include as first table segment some important dense BEV methods such as \texttt{BEVDet} \cite{BEVdet}, \texttt{BEVDepth} \cite{BEVDepth}, and \texttt{SOLOFusion} \cite{Solofusion}. All dense BEV methods employ the \texttt{ResNet-50 (RN-50)} \cite{resnet50} image encoder \cite{BEVdet,BEVDepth,Solofusion}. Note that the \texttt{ToC3D} results utilize the baseline layer, cf. Fig.\ \ref{fig:wmhsa_comparison}(a), while ours employ the proposed layer with hidden dimension $d_{h}=32$ in $\mathbf{TC}$, cf. Fig.\ \ref{fig:proposed_layer_tc}(b).

Noting that the different segments of Tab.\ \ref{table:quantitative_results} represent different regimes of model size, complexity, and latency, it does not come as a surprise that \texttt{RN-50}-based methods are outperformed by \texttt{SAM-B}-based ones, while \texttt{EVA-02-L}-based approaches are best for a given image resolution. Accordingly, let's focus on the \texttt{SAM-B} and \texttt{EVA-02-L} encoder regimes, which allow us to apply our efficient dynamic token selection approaches. 

In the \texttt{SAM-B} image encoder regime, the \texttt{StreamPETR} baseline delivers convincing mAP and NDS, with overall good other metrics as well. The \texttt{tgGBC} baseline lacks behind in these two important metrics, and also shows a worse average rank. Our methods (for $r\in\{0.5, 0.3\}$) have strongest average ranks in this regime while being significantly less complex and faster to compute than any of the two baselines.

In the \texttt{EVA-02-L} image encoder regime, again still for a $320 \times 800$ image resolution, \texttt{StreamPETR} shows good overall performance, along with an average rank of $2.90$. Interestingly, the two further baselines \texttt{tgGBC} and \texttt{ToC3D} fall behind with a shared average rank of $3.72$. \textit{In this regime, our method achieves best average ranks for all choices of $r$, whereby $r=0.1$ is both best in average rank, and by far the least complex and fastest method.} Specifically, compared to \texttt{ToC3D}, our approach with $r=0.1$
improves the mAP metric by $1.0\%$ absolute ($49.5\%$ vs.\ $50.5\%$) and NDS by $0.7\%$ absolute ($58.7\%$ vs.\ $59.4\%$), \textit{while reducing the GFLOPs by $55\%$ ($8,908$ vs.\ $4,030$) and the total latency by $25\%$ ($571$ ms vs.\ $430$ ms)}. This is achieved although model sizes in this regime are very similar among the methods. Comparing the second table segment to this third one, we observe that our method is able to be increasingly token-selective ($r \downarrow$) the larger the encoder gets---by that creating an attractive high-performance efficient multi-view 3D object detection.   

For the higher image resolution of $800 \times 1600$, among the baselines, \texttt{ToC3D} now leads the field with an average rank of $2.90$. We observe that our method for all values of $r$ is even better. \textit{Again, as with the lower image resolution, our most efficient approach ($r=0.1$) leads the field with a competitive average rank of $2.09$---again being best in computational complexity and latency.}

\begin{table}[t!]
  \caption{\textbf{Comparison of trainable and total model parameters} of our proposed approach vs.\ \texttt{ToC3D} on $\mathcal{D}_{\mathrm{NS}}^{\mathrm{val}}$ \cite{caesar2020nuscenes} with an \texttt{EVA-02-L} \cite{fang2024eva} encoder on three multi-view 3D object detection methods. Symbol $+$ indicates integration into the respective multi-view 3D object detection method. Best results bold.}
  \centering
  \setlength{\tabcolsep}{4pt} 
     \begin{tabular}{lrr}
          \toprule
          \textbf{Method} & \textbf{Fine-tuned (M)} & \textbf{Total (M)} \\
          \midrule 
          \texttt{StreamPETR} \cite{streampetr} & -- & $316.6$ \\
          $+$ \texttt{ToC3D} \cite{toc3D}     & $324.4$ & $+7.8$ \\
          $+$ Ours               & $\mathbf{1.6}$ & $+\mathbf{1.6}$ \\
          \midrule
          \midrule 
          \texttt{RayDN} \cite{raydn}       & -- & $331.9$ \\
          $+$ \texttt{ToC3D} \cite{toc3D}     & $339.6$ & $+7.7$  \\
          $+$ Ours               & $\mathbf{1.6}$ & $+\mathbf{1.6}$ \\
          \midrule
          \midrule 
          \texttt{Sparse4Dv2} \cite{sparse4dv2}  & --  & $316.5$ \\
          $+$ \texttt{ToC3D} \cite{toc3D}     & $324.1$ & $+7.6$ \\
          $+$ Ours               & $\mathbf{1.6}$ & $+\mathbf{1.6}$ \\
           \midrule 
          \bottomrule
      \end{tabular}
    \label{table:efficient_finetuning}
\end{table}

In Table \ref{table:efficient_finetuning}, we present a comparison of trainable and total model parameters of our proposed approach (Sections \ref{sec:proposed_layer_architecture} and \ref{sec:peft}) vs. \texttt{ToC3D} on $\mathcal{D}^{\mathrm{val}}_{\mathrm{NS}}$ using an \texttt{EVA-02-L} encoder \cite{fang2024eva}. Here, fine-tuned (trainable) parameters refer to the amount of model parameters that are updated during the (PEFT) fine-tuning process. Both methods are integrated into the \texttt{StreamPETR} \cite{streampetr}, \texttt{RayDN} \cite{raydn}, and \texttt{Sparse4Dv2} \cite{sparse4dv2} methods. Note that \texttt{ToC3D} \cite{toc3D} follows a full fine-tuning strategy of the image encoder $\mathbf{E}$, feature pyramid network $\mathbf{FPN}$, and the 3D object decoder $\mathbf{D}$ of the respective 3D object detection method. As shown in Table \ref{table:efficient_finetuning}, \textit{integrating \texttt{ToC3D} into any of the three methods requires to fine-tune more than $300$ M parameters, while we propose a parameter-efficient fine-tuning of only $1.6$ M}. Note that a large number of fine-tuned parameters significantly increases the training overhead, thereby reducing training efficiency \cite{hu2022lora,lei2025rethinking}. Further, the introduction of the learnable motion queries $\mathbf{Q}^{\mathrm{mot}}$ and the token scorer (cf. Fig.~\ref{fig:wmhsa_comparison}(a)) for image feature token selection in \texttt{ToC3D} increases the total number of parameters of the respective 3D object detection methods by about $7.6$ M. In our proposed layer (cf. Fig.\ \ref{fig:proposed_layer_tc}(a)), \textit{we employ a single token selection block and a light-weight token compensator for image feature token selection, with only $1.6$ M additional parameters.} 

In Table \ref{table:generalization_study}, we present a generalization study of our proposed method vs.\ \texttt{ToC3D} \cite{toc3D} employed within the two further multi-view 3D object detection methods \texttt{RayDN} \cite{raydn} and \texttt{Sparse4Dv2} \cite{sparse4dv2}. All evaluation metrics, GFLOPs, and latency are measured with an input resolution of $320 \times 800$ on $\mathcal{D}^{\mathrm{val}}_{\mathrm{NS}}$. Further, both methods employ an \texttt{EVA-02-L} encoder \cite{fang2024eva}. Note that our proposed layer's hidden dimension is $d_{h}=32$ in $\mathbf{TC}$, cf.\ Fig.\ \ref{fig:proposed_layer_tc}(b).


 Comparing Table \ref{table:quantitative_results} for the \texttt{EVA-02-L} encoder at input resolution $320 \times 800$ vs.\ Table \ref{table:generalization_study}, we observe at very similar model sizes that again \texttt{ToC3D} is poorest w.r.t.\ the average rank. The \texttt{RayDN} baseline method follows next, while in the lower table segment, the baseline \texttt{Sparse4Dv2} even achieves a good average rank of $2.54$. While our methods show average ranks between $2.27$ ... $2.63$, again, as with \texttt{StreamPETR} in Table \ref{table:quantitative_results}, also in the context of \texttt{RayDN} and \texttt{Sparse4Dv2} in Table \ref{table:generalization_study}, \textit{our approach with $r=0.1$ turns out to be the overall most attractive configuration: It reaches the top avg.\ rank of $2.27$ within \texttt{RayDN}}. 
 Compared to \texttt{ToC3D}, our approach with $r=0.1$ increases the mAP by $2.8\%$ absolute ($50.9\%$ vs.\ $53.7\%$) and NDS by $1.2\%$ absolute ($60.9\%$ vs.\ $62.1 \%$), \textit{while reducing the GFLOPs by $48\%$ ($9,359$ vs.\ $4,909$) and total latency by $9\%$ ($510$ ms vs.\ $464$ ms)}. 
 
 Also, within \texttt{Sparse4Dv2} (lower table segment), 
 our approach ($r=0.1$) attains an avg.\ rank of $2.54$, on par with the \texttt{Sparse4Dv2} baseline, while clearly outperforming \texttt{ToC3D} ($3.36$). Compared to \texttt{ToC3D}, our approach increases the mAP by $1.8\%$ absolute ($50.7\%$ vs.\ $52.5\%$) and NDS by $0.4\%$ absolute ($60.1 \%$ vs.\ $60.5\%$), \textit{while reducing the GFLOPs by $55\%$ ($8,682$ vs.\ $3,879$) and total latency by $10\%$ ($257$ ms vs.\ $230$ ms)}. Overall, these results demonstrate that our approach simultaneously improves both detection performance and efficiency across two additional multi-view 3D object detection methods.





\section{Conclusions}
\label{sec:conclusion}

In this work, we address a limitation of so-far state-of-the-art methods based on \texttt{ToC3D} for efficient multi-view 3D object detection, which is the fixed layer-individual token selection ratio. We introduce a dynamic layer-wise token selection strategy within large-scale pre-trained vision transformer backbones to accelerate existing 3D object detection methods. Further, instead of performing a computationally demanding full fine-tuning as in \texttt{ToC3D}, we propose a parameter-efficient fine-tuning (PEFT) strategy that trains only the proposed modules for dynamic token selection, while keeping the rest of the model weights frozen, thereby reducing the number of fine‑tuned parameters from more than $300$M to only $1.6$M. Across three multi-view 3D object detection approaches, we demonstrate that our proposed approach decreases the computational complexity (GFLOPs) by $48\%$ ... $55\%$, and inference latency (on an \texttt{NVIDIA-GV100} GPU) by $9\%$ ... $25\%$, while still improving the mAP by $1.0\%$ ... $2.8\%$ absolute and NDS by $0.4\%$ ... $1.2\%$ absolute compared to the so-far state of the art \texttt{ToC3D} on the NuScenes dataset.



\appendices

\section{Window-based Multi-head Self Attention (WMHSA)}
\label{app:wmhsa}

\begin{figure}[t!]
\centering
  \hspace{-5em}
  \resizebox{1.1\linewidth}{!}{\input{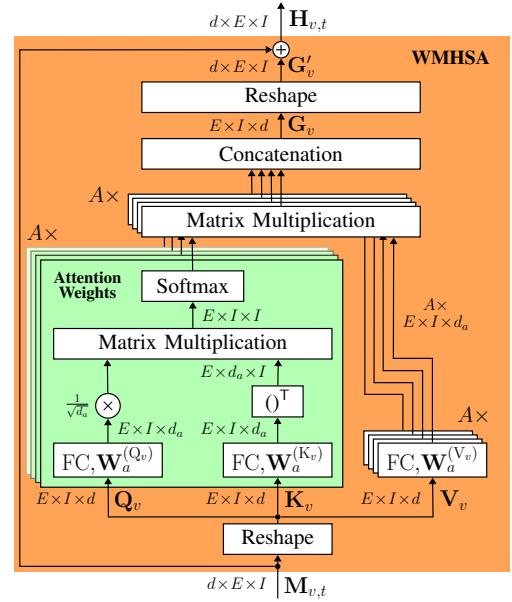}}
    \caption{\textbf{Window-based multi-head self attention (WMHSA)} \cite{liu2021swin} from Figs.\ \ref{fig:wmhsa_comparison}, and \ref{fig:proposed_layer_tc}. The internal dimension $d_{a}$ refers to the per-head internal dimension.   }
    \label{fig:attention_block}
\end{figure}

Fig.\ \ref{fig:attention_block} depicts the detailed architecture of the WMHSA block, employed by \texttt{SAM-B} \cite{SAM}, and \texttt{EVA-02-L} \cite{fang2024eva} image encoders. The input $\mathbf{M}_{v,t}$ is reshaped to produce the query $\mathbf{Q}_{v} \in \mathbb{R}^{E \times I \times d}$, key $\mathbf{K}_{v} \in \mathbb{R}^{E \times I \times d}$, and value $\mathbf{V}_{v} \in \mathbb{R}^{E \times I \times d}$, which are passed through the scaled dot-product self attention 
    \begin{equation}
            \mathbf{G}_{v,a} (\mathbf{Q}_{v},\mathbf{K}_{v},\mathbf{V}_{v}) = \operatorname{SM}(\tfrac{\mathbf{Q}_{v}\mathbf{W}_{a}^{(\mathrm{Q}_{v})} \cdot (\mathbf{K}_{v}\mathbf{W}_{a}^{(\mathrm{K}_{v})})^{\mathsf{T}}}{\sqrt{d}_{a}})\cdot\mathbf{V}_{v}\mathbf{W}_{a}^{(\mathrm{V}_{v})},
    \label{eq:wmhsa}
    \end{equation}
   to compute the window-based attention output $\mathbf{G}_{v,a} \in \mathbb{R}^{E \times I \times d_{a}}$ \textit{independently} across all $E$ windows, softmax operation $\operatorname{SM}$, learnable fully connected (FC) layers weights $\mathbf{W}_{a}^{(\mathrm{Q}_{v})}$, $\mathbf{W}_{a}^{(\mathrm{K}_{v})}$, $\mathbf{W}_{a}^{(\mathrm{V}_{v})} \in \mathbb{R}^{d \times d_{a}}$, and $()^{\mathsf{T}}$ being the transpose, applied only over the last two dimensions ($I \times d_a$) of each window. Here, $a \in \mathcal{A}$ refers to the head index, and $\mathcal{A} = \{1,\dots,A \}$ refers to the head index set, with $A \in \mathbb{N}$ being the number of heads. The result of the scaled dot-product self attention is concatenated to $\mathbf{G}_{v}=(\mathbf{G}_{v,a})\in \mathbb{R}^{E \times I \times d}$ and reshaped to produce $\mathbf{G}^{\prime}_{v} \in \mathbb{R}^{d \times E \times I}$, which is added to $\mathbf{M}_{v,t}$ to generate the refined representation $\mathbf{H}_{v,t} = (\mathbf{H}_{v,t,e}) = \mathbf{G}^{\prime}_{v} + \mathbf{M}_{v,t} \in \mathbb{R}^{d \times E \times I}$, with $\mathbf{H}_{v,t,e} \in \mathbb{R}^{d \times I}$, and respective window index $e \in \mathcal{E}$ from index set $\mathcal{E}=\{1,\dots,E\}$. Note that we dropped frame index $t$ for all WMSHA-internal entities for better readability. For the baseline layer, the refined representation $\mathbf{H}_{v,t}$ is employed to compute residual output $\mathbf{H}_{v,t}^{\prime}$. For the proposed layer, it is passed to the window unpartition block.

\bibliographystyle{IEEEtran}
\bibliography{references}


 




\vfill

\end{document}